\documentclass[sigconf]{acmart}

\AtBeginDocument{%
  \providecommand\BibTeX{{%
    \normalfont B\kern-0.5em{\scshape i\kern-0.25em b}\kern-0.8em\TeX}}}

\copyrightyear{2024} 
\acmYear{2024} 
\setcopyright{rightsretained} 
\acmConference[KDD '24]{Proceedings of the 30th ACM SIGKDD Conference on Knowledge Discovery and Data Mining}{August 25--29, 2024}{Barcelona, Spain}
\acmBooktitle{Proceedings of the 30th ACM SIGKDD Conference on Knowledge Discovery and Data Mining (KDD '24), August 25--29, 2024, Barcelona, Spain}
\acmDOI{10.1145/3637528.3671967}
\acmISBN{979-8-4007-0490-1/24/08}

\usepackage{hyperref}
\makeatletter
\def\UrlAlphabet{%
      \do\a\do\b\do\c\do\d\do\e\do\f\do\g\do\h\do\i\do\j%
      \do\k\do\l\do\m\do\n\do\o\do\p\do\q\do\r\do\s\do\t%
      \do\u\do\v\do\w\do\x\do\y\do\z\do\A\do\B\do\C\do\D%
      \do\E\do\F\do\G\do\H\do\I\do\J\do\K\do\L\do\M\do\N%
      \do\O\do\P\do\Q\do\R\do\S\do\T\do\U\do\V\do\W\do\X%
      \do\Y\do\Z}
\def\UrlDigits{\do\1\do\2\do\3\do\4\do\5\do\6\do\7\do\8\do\9\do\0}
\g@addto@macro{\UrlBreaks}{\UrlOrds}
\g@addto@macro{\UrlBreaks}{\UrlAlphabet}
\g@addto@macro{\UrlBreaks}{\UrlDigits}
\makeatother

\usepackage{multirow}
\usepackage{makecell}

\usepackage{caption}
\usepackage{subcaption} 
\usepackage{tabularx}
\usepackage{booktabs}
\usepackage{xspace}
\usepackage[utf8]{inputenc}
\usepackage{booktabs}
\usepackage{longtable}
\usepackage{graphicx}
\usepackage{tabularx}
\usepackage{booktabs}
\usepackage{ltablex}
\usepackage{geometry}
\usepackage{pdflscape}
\usepackage{booktabs}
\usepackage{adjustbox}
\usepackage{algorithm}
\usepackage{algorithmic}

\newcommand{\ours}[0]{\textsc{Magi}\xspace}

\begin{document}

\title{Revisiting Modularity Maximization for Graph Clustering: A Contrastive Learning Perspective}

\author{Yunfei Liu}
\authornote{Both authors contributed equally to this research.}
\affiliation{\institution{Ant Group}
  \country{}}
\email{leo.lyf@antgroup.com}

\author{Jintang Li}
\authornotemark[1]
\authornote{Corresponding author.}
\affiliation{\institution{Ant Group}
  \country{}}
\email{edisonleejt@gmail.com}

\author{Yuehe Chen}
\affiliation{\institution{Ant Group}
  \country{}}
\email{chenyuehe.cyh@antgroup.com}

\author{Ruofan Wu}
\affiliation{\institution{Ant Group}
  \country{}}
\email{wuruofan1989@gmail.com}

\author{Ericbk Wang}
\affiliation{\institution{Ant Group}
  \country{}}
\email{yike.wbk@antgroup.com}

\author{Jing Zhou}
\affiliation{\institution{Ant Group}
  \country{}}
\email{colin.zj@antgroup.com}

\author{Sheng Tian}
\affiliation{\institution{Ant Group}
  \country{}}
\email{tiansheng.ts@antgroup.com}

\author{Shuheng Shen}
\affiliation{\institution{Ant Group}
  \country{}}
\email{shuheng.ssh@antgroup.com}

\author{Xing Fu}
\affiliation{\institution{Ant Group}
  \country{}}
\email{fux008@gmail.com}

\author{Changhua Meng}
\affiliation{\institution{Ant Group}
  \country{}}
\email{changhua.mch@antgroup.com}

\author{Weiqiang Wang}
\affiliation{\institution{Ant Group}
  \country{}}
\email{wang.weiqiang@gmail.com}

\author{Liang Chen}
\affiliation{\institution{Unaffiliated}
  \country{}}
\email{jasonclx@gmail.com}

\renewcommand{\shortauthors}{Liu and Li, et al.}

\begin{abstract}

  Graph clustering, a fundamental and challenging task in graph mining, aims to classify nodes in a graph into several disjoint clusters. In recent years, graph contrastive learning (GCL) has emerged as a dominant line of research in graph clustering and advances the new state-of-the-art. However, GCL-based methods heavily rely on graph augmentations and contrastive schemes, which may potentially introduce challenges such as semantic drift and scalability issues. Another promising line of research involves the adoption of modularity maximization, a popular and effective measure for community detection, as the guiding principle for clustering tasks. Despite the recent progress, the underlying mechanism of modularity maximization is still not well understood. In this work, we dig into the hidden success of modularity maximization for graph clustering. Our analysis reveals the strong connections between modularity maximization and graph contrastive learning, where positive and negative examples are naturally defined by modularity. In light of our results, we propose a community-aware graph clustering framework, coined \ours, which leverages modularity maximization as a contrastive pretext task to effectively uncover the underlying information of communities in graphs, while avoiding the problem of semantic drift. Extensive experiments on multiple graph datasets verify the effectiveness of \ours in terms of scalability and clustering performance compared to state-of-the-art graph clustering methods. Notably, \ours easily scales a sufficiently large graph with 100M nodes while outperforming strong baselines.

\end{abstract}



\maketitle

\section{Introduction}\label{intro}
Graph clustering is a fundamental problem in graph analysis, crucial for uncovering structures and relationships between nodes in a graph. The primary objective of graph clustering is to group or partition the nodes in a graph into clusters or communities based on their structural properties or connectivity patterns. So far, graph clustering has been widely studied and extensively applied across various domains, including social network analysis~\cite{Newman2006Finding}, image segmentation~\cite{Felzenszwalb2004EfficientGI} and recommendation systems~\cite{Moradi2015effective}.

\begin{figure}[t]
  \centering
  \includegraphics[width=\linewidth]{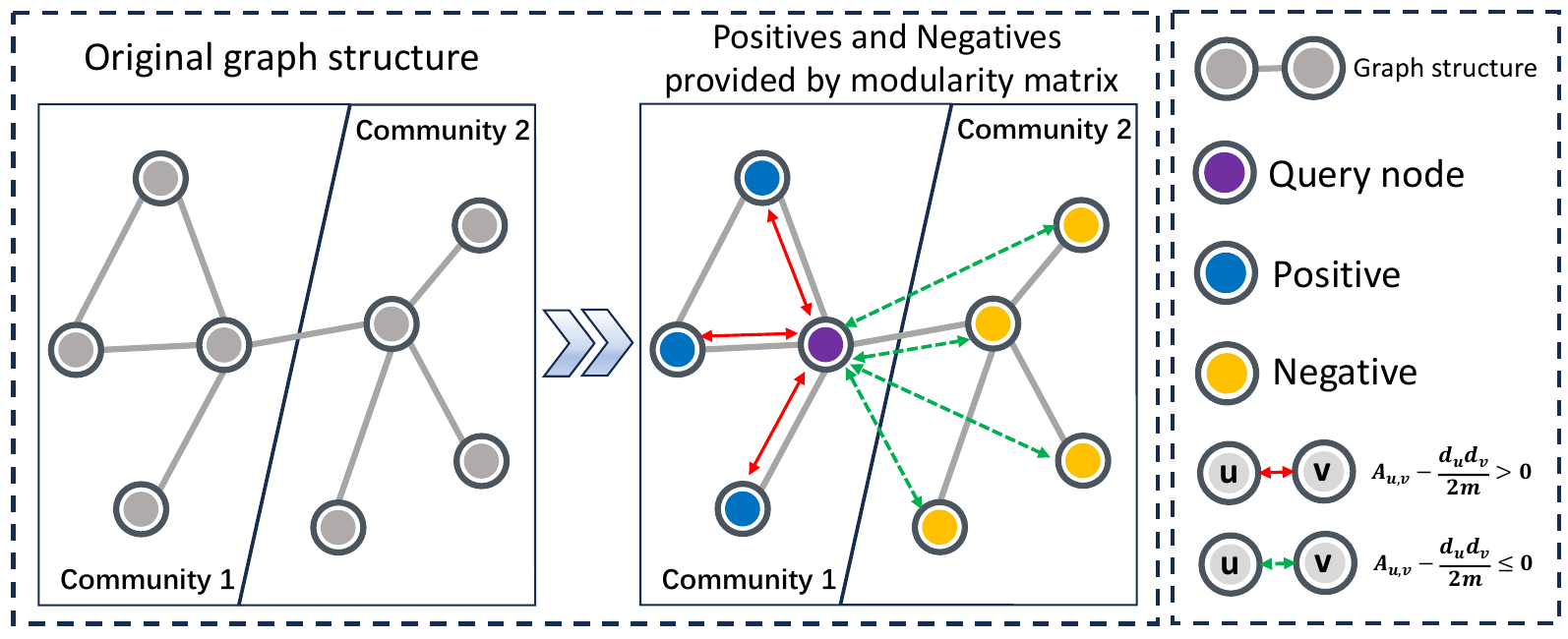}
  \caption{An illustrative overview of how positive and negative examples of a query node are guided by the ``modularity'' measure.}
  \label{fig:sketch}
  \vspace{-5mm}
\end{figure}

As a longstanding field of research, graph clustering continues to evolve through the development of novel algorithms and techniques. Recently, graph neural networks (GNNs) have emerged as the \textit{de facto} deep learning architecture for dealing with graph data~\cite{GCN, GAT}. In light of the learning capability of GNNs on graph-structured data, researchers have shifted their attention to exploring GNN-based approaches for graph clustering. Typically, GNNs are employed as encoders to learn node representations, which are often accomplished with an auxiliary task to help uncover the underlying patterns for clustering. As the graph clustering task is commonly approached in an unsupervised manner due to the absence of labeled annotations, this has motivated the exploration of self-supervised graph learning methods for graph clustering.
Contrastive learning~\cite{Hadsell2006Dimensionality, Wu2018CVPRUnsupervised}, which aims to learn representations that bring similar instances closer in the representation space and distances dissimilar ones, has been proven to learn ``cluster-preserving'' representations~\cite{Parulekar2023InfoNCELP}.
Therefore, recently, graph contrastive learning methods (GCLs)~\cite{DGI, spikegcl, GraphCL, SunLCWBZ24} have made significant breakthroughs in graph clustering and gradually become the mainstream approach for graph clustering.

Inspired by~\cite{Wu2018CVPRUnsupervised}, early GCLs~\cite{DGI, GRACE2020ICMLWorkshop} adopted instance discrimination as a pretext task, aiming to learn representations that are invariant to different augmented views of the graph. Recently, some works have proposed augmentation-free GCLs~\cite{AFGRL, CCGC2023AAAI, devvrit2022s3gc, SCGDN}, which extract contrastive information from the graph data itself and construct corresponding pseudo-labels. Several works~\cite{AFGRL, CCGC2023AAAI, SCGDN} use K-means~\cite{kmeans} and K-Nearest Neighbors (KNN) to mine the potential cluster within graph features, and utilize the clustering results as foundation for positive/negative sample pairs.
Although the GCLs mentioned above have achieved significant success in the graph clustering task, we note that these methods still suffer from at least one of the following challenges:

\textbf{(C1) Scalability.}
Mainstream GCLs typically rely on data augmentation to create multiple views and employ multiple encoders to obtain corresponding representations. However, this poses computational challenges and scalability issues when applied to large-scale graphs. On the other hand, feature-based augmentation-free GCLs require the use of high-cost algorithms such as K-means and KNN to construct pseudo-labels, which also limits their capability to scale to large-scale datasets.

\textbf{(C2) Semantic drift.}
Pretext tasks play an important role in enabling GCL to better adapt to downstream tasks.  A typical approach is to define instance-wise ``augmentations'' and then pose the problem as that of learning to closely align these augmentations with the original, while keeping them separate from others.
However, this type of pretext task neglects the inherent structure of graph data, which can result in semantic drift during downstream clustering tasks~\cite{AFGRL}.
Augmentation-free GCLs can mitigate this by mining the information inherently carried by graph data. However, as previously mentioned, methods based on K-means and KNN lack scalability, and using simple random walks to mine positive samples~\cite{devvrit2022s3gc} within a node's neighborhood can easily lead to a community semantic drift, especially when a node is situated at the edge of a community.

Recently, methods based on neural modularity~\cite{newman2006modularity} maximization have made new progress in graph clustering tasks~\cite{DMoN2023JMLR, DGCLUSTER2024AAAI}. By using a single GNN encoder to encode the relaxed community assignment matrix, these methods effectively combine the modularity maximization objective with GNNs and achieve state-of-the-art performance in graph clustering tasks. The modularity maximization objective can effectively perceive the potential community structure within networks~\cite{Liu2022Robust} and provide guidance for representation learning.
However, the design of the modularity function is heuristic, and the underlying reasons for its success as an optimization objective remain largely unexplored.

In this work, we provide an in-depth analysis of modularity maximization and bridge the gap between modularity maximization and GCL. Our analysis shows that modularity maximization is essentially graph contrastive learning, where the positive and negative examples are naturally guided by the modularity matrix (see Figure~\ref{fig:sketch}). Based on our findings, we attempt to integrate the latest developments in the fields of neural modularity maximization and graph contrastive learning, and propose a co\underline{m}munity-\underline{a}ware \underline{g}raph cluster\underline{i}ng framework, coined \ours.
\ours can mitigate the effects of semantic drift by perceiving the underlying community structures in the graph, and since it doesn't rely on data augmentation, it can easily scale to a sufficiently large graph with 100M nodes. Our contributions can be summarized as follows:

\begin{itemize}
  \item \textbf{Modularity maximization $=$ contrastive learning.} We establish the connection between modularity maximization and graph contrastive learning. Our findings reveal that modularity maximization can be viewed as leveraging potential community information in graphs for contrastive learning.

  \item \textbf{Community-aware pretext task and scalable framework.} We propose \ours, a community-aware graph contrastive learning framework that uses \textit{modularity maximization} as its pretext task. \ours avoids semantic drift by leveraging underlying community structures and eliminates the need for graph augmentation. \ours incorporates a two-stage random walk approach to perform modularity maximization pretext tasks in mini-batch form, thereby achieving good scalability.

  \item \textbf{Experimental results.} We conduct extensive experiments on 8 real-world graph datasets with different scales. \ours has consistently outperformed several state-of-the-arts in the task of graph clustering. Notably, \ours easily scales to an industrial-scale graph with 100M nodes, showcasing its scalability and effectiveness in large-scale scenarios.
\end{itemize}

\section{Related Work}
\subsection{Graph clustering}
Graph clustering is a widely studied problem in academia and industry. Classical clustering methods involve either solving an optimization problem or using some heuristic, non-parametric approaches. Prominent examples include K-means~\cite{kmeans}, spectral clustering~\cite{Jianbo2000Normalized}, and Louvain~\cite{louvain}. However, the shallow architecture of these methods limits their performance.
With the rise of deep neural networks in graph representation learning, random walk-based methods such as DeepWalk~\cite{deepwalk} and Node2vec~\cite{node2vec} have also been introduced for addressing clustering tasks.

Early works typically focus on single dimension form graphs, e.g., graph structure or node attributes. Thanks to the great ability of graph neural networks (GNNs)~\cite{GCN, GAT} in learning jointly structure and attribute information, graph autoencoders stand out as an emerging approach for unsupervised graph learning tasks. For example, GAE and VGAE~\cite{VGAE} learn to reconstruct the graph structure as the self-supervised learning task using GNNs. Follow-up works extend GAE by employing Laplacian sharpening~\cite{GALA}, Laplacian smoothing~\cite{AGE2020KDD}, generative adversarial learning~\cite{ARVGA}, and masked autoencoding~\cite{maskgae}. However, self-supervised learning tasks or pretext tasks in GAEs are not aligned with downstream tasks such as graph clustering. As a result, the learned representations may not effectively capture the relevant information for clustering tasks.

\subsection{Graph contrastive learning}
Over the past few years, graph contrastive learning (GCL) has emerged as a powerful technique for learning representations of graph-structured data. Deep Graph Infomax (DGI)~\cite{DGI} follows the approach of mutual information-based learning, as proposed by
adapted~\cite{hjelm2018learning}. GRACE~\cite{GRACE2020ICMLWorkshop} maximizes the agreement of node representations between two corrupted views of a graph. GraphCL~\cite{GraphCL} incorporates four graph augmentation techniques to learn unsupervised graph representations through a unified contrastive learning framework. One step further, MVGRL~\cite{MVGRL2020ICML} introduces node diffusion and contrasts node-level embeddings with representations of augmented graphs. Following the BYOL~\cite{BYOL}, BGRL~\cite{BGRL2021ICLRWorkshop} eliminates the need for negative sampling by minimizing an invariance-based loss for augmented graphs within a batch. In light of the success of GCL, there have been attempts to apply GCL techniques to graph clustering and achieve promising results~\cite{AFGRL,SCGDN,CCGC2023AAAI,devvrit2022s3gc}.
However, most methods employ classical instance discrimination as the pretext task, which requires sophisticated graph augmentation techniques to obtain meaningful view representations. As a result, they may suffer from challenges such as semantic drift and limitations in scaling up to handle large graphs.

\subsection{Neural modularity maximization}
As one of the most commonly used metrics for community detection, modularity measures the quality of a partition in a network by evaluating the density of connections within communities compared to random connections~\cite{newman2006modularity}.
Maximizing the modularity directly is proven to be NP-hard~\cite{Brandes2006MaximizingMI}.
As a result, heuristics such as spectral relaxation~\cite{Newman2006Finding} and greedy algorithms~\cite{louvain} have been developed to approximate the modularity and find suboptimal but reasonable community partitions.
However, previous works focus simply on the graph structure, while ignoring the abundant information associated with the nodes (e.g., attributes). With the advent of graph neural networks, combining them with modularity maximization has become a promising research direction.
\cite{Yang2016Modularity, Bhatia2018DFuzzyAD} utilizes an autoencoder to encode the community assignment matrix and combines it with a reconstruction loss to train the autoencoder. \cite{Sun2020Network, Liu2022Robust} and \cite{Choong2018Learning, Guillaume2022Modularity, Zhou2022End} extend this idea by employing (variational) graph autoencoders. \cite{Liu2022Robust} explored the benefits brought by high-order proximity and used high-order polynomials of the adjacency matrix to calculate the high-order modularity matrix. \cite{Sun2022Graph} constructs two loss functions based on modularity, corresponding to single and multi-attribute networks, while DMoN~\cite{DMoN2023JMLR} introduces collapse regularization to prevent the community allocation matrix from falling into spurious local minima.
Despite the recent progress, the underlying mechanism of neural modularity maximization for graph clustering is still not well understood.

\section{Motivation}
In this section, we first provide basic notations throughout this paper. We then introduce preliminary knowledge for modularity-based learning, which is connected to contrastive learning as our motivation.

\subsection{Problem statement and notations}
Given an attribute graph $\mathcal{G} = (\mathcal{V}, \mathcal{E}, \mathbf{X})$, where $\mathcal{V} = \{v_1, v_2,..., v_N\}$ denotes a set of nodes and $N=|\mathcal{V}|$; $\mathcal{E} \subseteq \mathcal{V} \times \mathcal{V}$ denotes corresponding edges between nodes, where each node $v \in \mathcal{V}$ is associated with a $d_h$-dimensional feature vector $x_v \in \mathbb{R}^{d_h}$. Let $\mathbf{X} \in \mathbb{R}^{N \times d_h}$ denote the node attribute matrix and $\mathbf{A} \in \mathbb{R}^{N \times N}$ the adjacency matrix, respectively. Given the graph $\mathcal{G}$ and node attributes $\mathbf{X}$, the goal of graph clustering is to partition the nodes set $\mathcal{V}$ into $C$ partitions $\{\mathcal{V}_1, \mathcal{V}_2, \mathcal{V}_3, \ldots, \mathcal{V}_C\}$ such that nodes in the same cluster are similar/close to each other in terms of the graph structure as well as in terms of attributes.

\subsection{Revisiting modularity maximization}
Modularity~\cite{newman2006modularity} is a measure of the graph structure, which measures the divergence between the intra-community edges
from the expected one. Formally, given a graph $\mathcal{G}$, the modularity~($Q$) is defined as:
\begin{equation} \label{eq:modulairty}
  Q = \frac{1}{2m} \sum_{i,j} (\mathbf{A}_{i,j} - \frac{d_i d_j}{2m})\delta(c_i, c_j),
\end{equation}
where $d_i = \sum_{j}\mathbf{A}_{i, j}$ is the degree of node $v_i$, $m =|\mathcal{E}|$ is the total number of edges in the graph, and $c_i$ is the community to which node $v_i$ is assigned.
$\delta(c_i, c_j) = 1$ is an indicator function, i.e., $\delta(c_i, c_j) = 1$ if $c_i=c_j$ and 0 otherwise, suggesting whether nodes $v_i$ and $v_j$ belong to the same community. Essentially, $\mathbf{A}_{i, j}$ and $\frac{d_id_j}{2m}$ can be regard as the ``observed'' and ``expected'' number of edges between nodes $v_i$ and $v_j$.
Since a larger ($Q$) leads to a better community partition, modularity maximization has become a popular approach in finding good community division~\cite{Newman2006Finding}.

Typically, modularity maximization can be viewed as a constrained optimization objective in the following form:
\begin{equation} \label{eq:modulairty_matrix_form}
  \begin{split}
    & \mathop{\max}_{\mathbf{P}} \quad  Q := tr(\mathbf{P}^{T} \mathbf{B} \mathbf{P}), \\
    & \ \ s.t. \  \quad \mathbf{P} \in \{0,1\}^{N \times C}, \\
    & \qquad \quad  tr(\mathbf{P}^{T} \mathbf{P}) = N,
  \end{split}
\end{equation}
where $\mathbf{B} \in \mathcal{R}^{N \times N}$ is the \textit{modularity matrix}, with each element $\mathbf{B}_{i,j} = \mathbf{A}_{i,j} - \frac{d_id_j}{2m}$ the \textit{modularity coefficient}. $\mathbf{P} =\{p_1, p_2, \ldots, p_N\}\in \{0,1\}^{N \times C}$ is a one-hot matrix indicating the community ownership of each node in the graph, i.e., \textit{community assignment matrix}. Typically, $\mathbf{P}_{i, j} = 1$ if $v_i$ belongs to community $\mathcal{V}_j$ and 0 otherwise.

\subsection{Connecting modularity maximization to contrastive learning}
Based on the preliminary knowledge introduced above, we now provide our insights to demonstrate the underlying relationship between modularity maximization and contrastive learning. Formally, we rewrite Eq.~\eqref{eq:modulairty} into an equivalent form below:
\begin{equation}
  \label{eq_modularity_contrastive}
  \begin{split}
    \mathop{\max}_{\mathbf{P}} \quad  Q &:= tr(\mathbf{P}^{T} \mathbf{B} \mathbf{P}) = \frac{1}{2m}\sum_{i,j} \mathbf{B}_{i,j} p_i^T p_j \\
    & = \frac{1}{2m} \left(\sum_{(v_i,v_j) \in \mathcal{M}^{+}} \mathbf{B}_{i,j} p_i^T p_j \ +  \sum_{(v_i,v_j) \in \mathcal{M}^{-}} \mathbf{B}_{i,j} p_i^T p_j\right) \\
    s.t. \quad \quad  &  p_i \in \{0,1\}^{C} \ \text{and} \ p_i^Tp_i = 1 \ \forall \ i=1,2,...,N,
  \end{split}
\end{equation}
where $\mathcal{M}^{+}=\{(v_i, v_j)| \mathbf{B}_{i,j}>0\}$ and $\mathcal{M}^{-}=\{(v_i, v_j)| \mathbf{B}_{i,j} \leq 0\}$ are positive and negative pairs, respectively. In this regard, modularity maximization is similar in form to graph autoencoders, with the goal of optimization is to reconstruct the modularity matrix $\mathbf{B}$ rather than adjacency matrix $\mathbf{A}$. Motivated by a recent work~\cite{maskgae} that showed the equivalence between graph autoencoders and contrastive learning, we provide our intuition to explicitly relate modularity maximization to contrastive learning.

Here we further unify Eq.~\eqref{eq_modularity_contrastive} into a general form from the perspective of optimization.
Since directly maximizing Eq.~\eqref{eq_modularity_contrastive} is intractable, we can now relax $\mathbf{P} \in \{0,1\}^{N \times C}$ to its continuous analog $\mathbf{Z} \in \mathbb{R}^{N \times C}$, where $\mathbf{Z}$ is node representations learned from a graph encoder $f$ such that $\mathbf{Z}=f(\mathcal{G})$.
\begin{equation}\label{eq:cl_form}
  \small
  \begin{split}
    \mathop{\max}_{f,g} \quad Q &= \frac{1}{2m}\left(\sum_{(v_i,v_j) \in \mathcal{M}^{+}}  \mathbf{B}_{i,j}z_i^T z_j \ +  \sum_{(v_i,v_j) \in \mathcal{M}^{-}} \mathbf{B}_{i,j}z_i^T z_j\right) \\
    &= \frac{1}{2m}\left(\sum_{(v_i,v_j) \in \mathcal{M}^{+}}  |\mathbf{B}_{i,j}|z_i^T z_j \ -  \sum_{(v_i,v_j) \in \mathcal{M}^{-}} |\mathbf{B}_{i,j}|z_i^T z_j\right) \\
    &= \frac{1}{2m} \left(\sum_{(v_i,v_j) \in \mathcal{M}^{+}}  g(z_i,z_j) \ -  \sum_{(v_i,v_j) \in \mathcal{M}^{-}} g(z_i,z_j)\right),
  \end{split}
\end{equation}
where $g(\cdot,\cdot)$ is the decoder network that decodes the pairwise node representations $(z_i, z_j)$ into the modularity score. Typically, the decoder $g$ can be simply defined by cosine similarity function, i.e., $g(z_i, z_j)=z_i^{T}z_j$ or a neural network.
Community coefficient $\mathbf{B}_{i,j}$ is absorbed into $g$ and $f$ and should be adjusted by the neural network during training. According to Eq.~\eqref{eq:cl_form}, we obtain a general form of graph contrastive learning, where positive and negative pairs are guided by the modularity coefficient $\mathbf{B}_{i,j}$.

\subsection{Opportunity: community-aware pretext task}

As discussed in Section~\ref{intro}, current GCL methods for graph clustering, whether augmentation-based or augmentation-free, may face challenges related to scalability and/or semantic drift. Therefore, an alternative yet promising way is to utilize modularity maximization as a pretext task to generate community-aware pseudo-labels that guide contrastive learning.
Specifically, the employment of the modularity matrix allows us to reveal the underlying community structure within the graph. In this regard, positive sample pairs are naturally defined as connected node pairs within the same community, while negative sample pairs are unconnected node pairs from different communities. This leverages the intrinsic community structure in the network to mitigate the impact of semantic drift while also eliminating the need for graph augmentations that may hinder its scalability.
Moreover, by establishing the relationship between modularity maximization and contrastive learning, we are able to integrate the latest advancements in modularity maximization into research on graph contrastive learning. Recent advances in the field of modularity maximization mainly include: (1) neural modularity maximization~\cite{DGCLUSTER2024AAAI, DMoN2023JMLR}; (2) high-order proximity in modularity~\cite{Liu2022Robust}; (3) Better parallel optimization algorithms~\cite{Leiden}, etc. The ones that are closely related to contrastive learning are (1) and (2). In this work, we propose to obtain cluster assignment via GNN encoder~\cite{GCN, GraphSAGE}, which allows (soft) cluster assignment to be differentiable for neural modularity maximization. In addition, we fully consider high-order proximity in modularity, and propose a sampling method based on random walk to ensure scalability while capturing high-order proximity within the community.

\section{Method}

In this section, we present an overview of the architecture of \ours, which comprises a graph convolutional encoder, a two-stage random walk sampler that serves the modularity maximization pretext task, and a SimCLR~\cite{simclr} contrastive loss. Each of these components will be introduced separately in the following subsections.
\begin{figure*}[t]
  \centering
  \includegraphics[width=\linewidth]{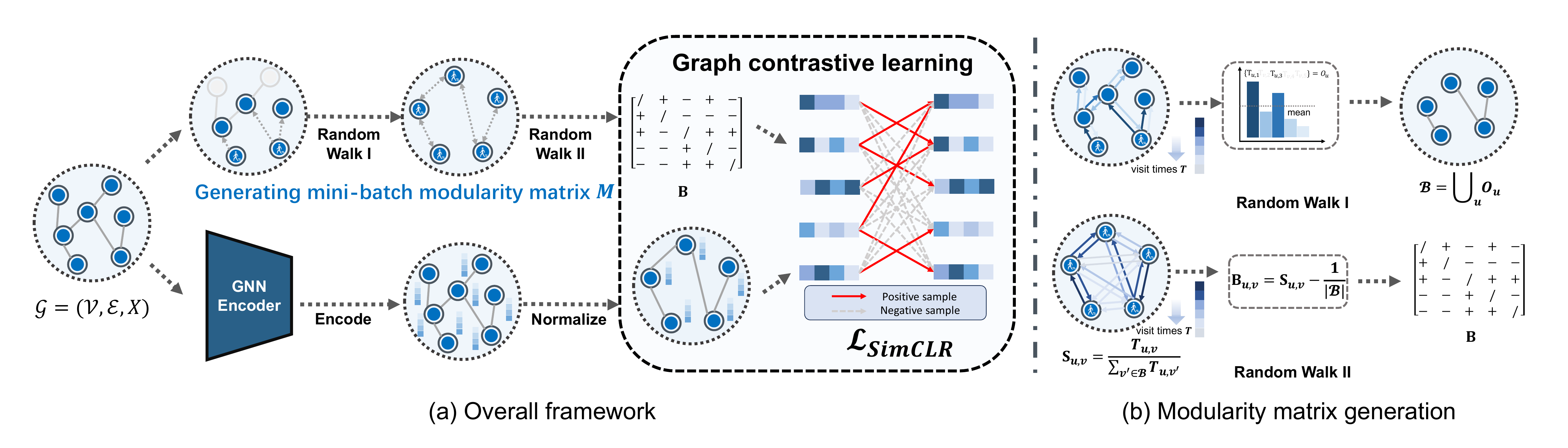}
  \caption{Overview of proposed \ours framework. During the self-supervised learning phase, the original graph $\mathcal{G}$ is provided to generate node embedding and modularity matrix through a GNN encoder and a two-stage random walk process, respectively. Then, the SimCLR loss function is employed to optimize the encoder in a self-supervised manner.}
  \label{fig:framework}
\end{figure*}

\subsection{GNN encoder}

An encoder in GCL is a crucial component that maps the input graph data into latent representations. In \ours, we employ different GNNs as encoders to encode graphs of varying sizes. Here we consider only two representative GNNs, i.e., GCN~\cite{GCN} and GraphSAGE~\cite{GraphSAGE}. It is important to note that \ours is not specific to any particular GNN model and can be applied with other architectures (e.g., GAT~\cite{GAT}) as well.

\textbf{GCN.}
GCN is a popular GNN model that has been widely used for various graph-based tasks. GCN operates by aggregating information from neighboring nodes and propagating it through multiple layers. It leverages the spectral graph convolution operation, which is based on the graph Laplacian matrix, to capture local and global graph structure. Formally, the message propagation of $l$-th layer in GCN is defined as:
\begin{equation}
  \label{GCN propagation}
  h_u^{(l)}=\operatorname{LeakyReLU}\left(\sum_{v \in \mathcal{N}_u \cup\{u\}} \frac{\mathbf{W}^{(l)} h_v^{(l-1)}}{\sqrt{\left(\left|\mathcal{N}_u\right|+1\right)\left(\left|\mathcal{N}_v\right|+1\right)}}\right),
\end{equation}
where $\operatorname{LeakyReLU}(\cdot) = max(0, \cdot) + \alpha * min(0, \cdot)$ is activation function. $\mathcal{N}_u$ is the set of adjacent neighbors of node $u$. $\mathbf{W}$ is the learnable weight of layer $l$ and $h^{(l)}$ is the latent representation with $h^{(0)}=x$.

\textbf{GraphSAGE.}
GCN relies on a symmetric and fixed aggregation function based on the graph Laplacian, which can limit its ability to capture diverse neighborhood structures and scale to large graphs effectively.
GraphSAGE is a GNN alternative designed to better handle large-scale graphs in an inductive manner:
\begin{equation}
  \label{GraphSAGE propagation}
  h_u^{(l)}=\operatorname{LeakyReLU}\left(\mathbf{W}^{(l)}_1 h_u^{(l-1)}+\mathbf{W}^{(l)}_2 \Theta(\{h_v^{(l-1)}| v \in \mathcal{N}_u\})\right),
\end{equation}
where $\Theta$ denotes different aggregation functions, such as mean, max, or LSTM-based aggregation. This allows GraphSAGE to capture diverse types of neighborhood information, accommodating various graph structures and improving its capability to learn from different node interactions.

\subsection{Modularity maximization pretext task}
The core insight behind neural modularity maximization is the computation of the modularity matrix $\mathbf{B}$. In order to adapt the minibatch training to obtain good scalability, an intuitive idea is to randomly sample $n$ nodes in the graph and obtain contrastive pairs from their corresponding sub-modularity matrix. However, it is important to note that this approach potentially introduces the following issues:
\begin{itemize}
  \item \textbf{Structure bias.} The sampled subgraph may not necessarily share a similar structure with the overall graph, which can hinder the performance of contrastive learning. Nevertheless, this issue becomes more serious as the graph becomes larger and sparser.
  \item \textbf{Lack of high-order proximity.} The vanilla modularity matrix only focuses on the first-order proximity within the graph, which contradicts our intuitive understanding of community structure in networks, that is, two nodes often belong to the same community due to high-order proximity, such as having many common neighbors. Although a recent work~\cite{Liu2022Robust} explored the benefits brought by high-order proximity and used high-order polynomials of the adjacency matrix $\mathbf{A}$ to calculate the high-order modularity matrix. However, the cost of computing high-order polynomials of the adjacency matrix $\mathbf{A}$ for large-scale graphs is still prohibitive. We discuss the benefits of incorporating high-order proximity through experimental investigation in Section~\ref{SemanticDriftExperiment}.
\end{itemize}

In this work, we propose a two-stage random walk-based sampling method to address the aforementioned challenges, as by adjusting the depth of the random walks, we can effectively capture the high-order proximity within the network. Specifically, we employ a first-stage random walk to sample multiple sub-communities and merge them into a training batch, ensuring that the corresponding subgraph within the batch contains effective community structures to guide the model. Subsequently, we perform a second-stage random walk to generate a similarity matrix between nodes within the batch, and we draw on the concept of the configuration model~\cite{Brede2012NetworksAnIM} to calculate the corresponding modularity matrix in the mini-batch form.

Next, we will provide a detailed description of these two stages, which we refer to as \textbf{S1} and \textbf{S2}, respectively.

\textbf{(S1) Sampling multiple sub-communities.}
For any $v$ in set $\mathcal{N}$ consisting of $n$ randomly selected root nodes in the graph, based on the principle of internal connectivity within communities, there must be an overlap between the neighborhood of node $v$ and its potential community, denoted as $\mathcal{O}_{v}$. Accurately selecting $\mathcal{O}_{v}$ can be an expensive task because it involves detecting potential community structures in the graph. However, based on the definition that \textit{a community is a set of nodes is densely connected internally}, we utilize random walks to approximate $\mathcal{O}_{v}$. To be specific, we initiate $t$ random walks of depth $l$ rooted at node $v$ and denote the set of visited nodes as $\mathcal{U}_{v}^{1}$, with the corresponding visit count recorded as $T_{v}^1$, which each element $t_{v, u}^1$ as the number of visits from $v$ to $u$. We filter $\mathcal{O}_{v}$ in the following way:
\begin{equation}\label{eq:stage_one_filter}
  \mathcal{O}_{v} = \{u \in \mathcal{U}_{v}^1 \ | \ t_{v, u}^1 > \frac{\sum_{u^{\prime}} t_{v, u^{\prime}}^1}{|\mathcal{U}_{v}^{1}|}\}
\end{equation}
In brief, we consider nodes with visits greater than the mean as $\mathcal{O}_{v}$. Then, for nodes in $\mathcal{N}$, we perform the same operation and set $\mathcal{B} = \cup_{v \in \mathcal{N}} \mathcal{O}_{v}$ as a training batch.

\textbf{(S2) Mini-batch modularity matrix.}
It is intuitive to treat pair nodes from the same sub-community as positives and those from different sub-communities as negatives. However, we must consider the potential overlaps between different sub-community, necessitating further exploration of the relationships among nodes in $\mathcal{B}$. Specifically, we perform again $t$ random walks of depth $l$ for each node in $\mathcal{B}$. For any two nodes $v$ and $u$ in $\mathcal{B}$, let $t_{v, u}^2$ be the number of visits from $v$ to $u$, we can then construct a similarity matrix $\mathbf{S}$ for the nodes in $\mathcal{B}$ based on the number of visits, which each element:
\begin{equation}\label{eq:similarity_matrix}
  \mathbf{S}_{v, u} = \frac{t_{v, u}^2}{\sum_{u^{\prime} \in \mathcal{B}} t_{v, u^{\prime}}^2}
\end{equation}
denotes probability that node $v$ is connected to node $u$.
Subsequently, following~\cite{newman2006modularity} we employ the configuration model~\cite{Brede2012NetworksAnIM} to estimate the expected probability of connection between any two nodes in $\mathcal{B}$. Ultimately, we get the mini-batch form of modularity matrix $\mathbf{B}$, where each element:
\begin{equation}\label{eq:minibatch_modularity_matrix}
  \mathbf{B}_{v, u} = \mathbf{S}_{v, u} - \frac{\sum_{u^{\prime} \in \mathcal{B}}\mathbf{S}_{v, u^{\prime}} \sum_{v^{\prime} \in \mathcal{B}} \mathbf{S}_{v^{\prime}, u}}{\sum_{v^{\prime},u^{\prime}} \mathbf{S}_{v^{\prime}, u^{\prime}}} = \mathbf{S}_{v, u} - \frac{1}{|\mathcal{B}|}
\end{equation}

\subsection{Contrastive loss formulation}
Given a train batch $\mathcal{B}$ and corresponding modularity matrix $\mathbf{B}$, for any nodes $v \in \mathcal{B}$, set $\mathcal{M}_{v}^{+} = \{u \in \mathcal{B} | \mathbf{B}_{v, u} > 0\}$ and $\mathcal{M}_{v}^{-} = \{u \in \mathcal{B} | \mathbf{B}_{v, u} <= 0\}$ are the positive and negative samples of $v$, respectively. Now a simple contrastive loss used in Eq.(\ref{eq:modulairty}) as:
\begin{equation}\label{eq:cl_loss_simple}
  \mathcal{L}_{simple}(v) = - \sum_{u \in \mathcal{M}^{+}}  sim(z_v, z_u) \  +  \sum_{u^{\prime} \in \mathcal{M}^{-}} sim(z_v, z_{u^{\prime}}),
\end{equation}
where $sim$ is inner product similarity function: $sim(z_v, z_u) = \frac{z_v^T z_u}{\|z_v\|\|z_u\|}$.

However, the above loss function demonstrates significantly inferior performance~\cite{Wang2021Understanding} compared to the softmax-based SimCLR contrastive loss~\cite{simclr}, which formulated as:
\begin{equation}\label{eq:SimCLR_loss}
  \scriptsize
  \mathcal{L}_{SimCLR}(v) = - \sum_{u \in \mathcal{M}_v^{+}} \log \frac{exp(z_v \cdot z_u / \tau)}{\sum_{u \in \mathcal{M}_v^{+}} exp(z_v \cdot z_u / \tau) + \sum_{u^{\prime} \in \mathcal{M}_v^{-}} exp(z_v \cdot z_{u^{\prime}} / \tau)}
\end{equation}
The softmax-based SimCLR contrastive loss is a hardness-aware loss function~\cite{Wang2021Understanding}, where $\tau$ is temperature and plays a key role in controlling the local separation and global uniformity of the representation distributions. Appropriate temperature selection can effectively alleviate the uniform-tolerance dilemma.
Based on the above analysis, we use the SimCLR loss function defined in Eq.(\ref{eq:SimCLR_loss}) to learn the encoder.

\subsection{Complexity analysis}
The space complexity of our algorithm is $\mathcal{O}(Nd + m + b^2)$, where $N$ and $m$ are the number of nodes and edges in the graph, respectively. $d$ is the attribute dimension, and $b$ is the batch size. For all datasets, we store the graph feature $\mathbf{X}$, the sparse adjacency matrix $\mathbf{A}$, and the mini-batch modularity matrix $\mathbf{B}$ in memory, which require $\mathcal{O}(Nd)$, $\mathcal{O}(m)$, and $\mathcal{O}(b^2)$ storage space, respectively. For GPUs, since only the subgraph data corresponding to each batch needs to be stored, the space complexity is reduced to $\mathcal{O}(b(b + d))$. For a given batch, the forward and backward computation costs $\mathcal{O}(b d^2)$. Hence, for $N$ nodes, and $r$ epochs, time complexity is $\mathcal{O}(Nrd^2)$. We compare the time and space complexity of all methods used in Table~\ref{table:main_clustering_resluts}, and the result (see Table~\ref{table:complexity} in Appendix) shows that \ours performs better in terms of both time and memory complexity as compared to other methods that leverage both graph and feature information.

\section{EXPERIMENTS}
In this section, we conducted extensive experiments on several node classification benchmark datasets to evaluate the performance of our method in comparison with state-of-the-art baselines. Code is made publicly available at \url{https://github.com/EdisonLeeeee/MAGI}.

\subsection{Datasets}
In our experiments, we employ four small scale datasets (Cora~\cite{GCN}, Citeseer~\cite{GCN}, Photo~\cite{Oleksandr2018Pitfalls}, Computers~\cite{Oleksandr2018Pitfalls}), three large scale datasets (Reddit~\cite{GraphSAGE}, ogbn-arxiv~\cite{hu2020ogb}, ogbn-products~\cite{hu2020ogb}) and one extra-large scale dataset (ogbn-papers100M~\cite{hu2020ogb}) to showcase the effectiveness of our method. The details of the datasets are presented in Table ~\ref{dataset_stats}.

\begin{table}[t]
  \caption{Dataset statistics.}
  \label{dataset_stats}
  \begin{adjustbox}{width=0.5\textwidth,center}
    \large
    \centering
    \begin{tabular}{l|cccc}
      \toprule
      \textbf{Dataset} & \textbf{\# Nodes} & \textbf{\# Edges} & \textbf{\# Features} & \textbf{\# Clusters} \\
      \addlinespace
      \midrule
      \midrule
      \addlinespace
      Cora             & 2,708             & 5,278             & 1,433                & 7                    \\
      CiteSeer         & 3,327             & 4,614             & 3,703                & 6                    \\
      Amazon-Photo     & 7,650             & 119,081           & 745                  & 8                    \\
      Amazon-Computers & 13,752            & 491,722           & 767                  & 10                   \\
      ogbn-arxiv       & 169,343           & 1,166,243         & 128                  & 40                   \\
      Reddit           & 232,965           & 23,213,838        & 602                  & 41                   \\
      ogbn-products    & 2,449,029         & 61,859,140        & 100                  & 47                   \\
      ogbn-papers100M  & 111,059,956       & 1,615,685,872     & 128                  & 172                  \\
      \bottomrule
    \end{tabular}
  \end{adjustbox}
\end{table}

\subsection{Baselines}\label{Baseline}
We compare \ours with 11 baselines, which can be categorized into five groups:

\textbf{(1) Methods utilize graph structure only.} Node2vec~\cite{node2vec} is a scalable graph embedding technique that utilizes random walks on the graph structure.

\textbf{(2) Methods utilize graph features only.} K-means\cite{kmeans} is a traditional clustering algorithms that utilize only graph features.

\textbf{(3) GCLs with augmentations.} DGI~\cite{DGI}, GRACE~\cite{GRACE2020ICMLWorkshop}, MVGRL~\cite{MVGRL2020ICML} and BGRL~\cite{BGRL2021ICLRWorkshop} are GCLs based on data augmentation to learn node representations.

\textbf{(4) Augmentation-free GCLs.} CCGC~\cite{CCGC2023AAAI}, SCGDN~\cite{SCGDN}, and S$^{3}$GC~\cite{devvrit2022s3gc} are recent augmentation-free GCLs.

\textbf{(5) Methods based on neural modularity maximization.} DGCLUSTER~\cite{DGCLUSTER2024AAAI} and DMoN~\cite{DMoN2023JMLR} are two recent neural modularity maximization methods that can achieve good performance in graph clustering, in which DGCLUSTER is a semi-supervised method.

\subsection{Metrics}
Following the evaluation setup of \cite{MVGRL2020ICML, devvrit2022s3gc}, we measure 4 metrics related to evaluating the quality of cluster assignments: Accuracy (ACC), Normalized Mutual Information (NMI), Adjusted Rand Index (ARI) and Macro-F1 Score (F1). For all the aforementioned metrics, higher values indicate better clustering performance. In our experiments, we first generate representations for each method and then perform spectral clustering on the embeddings of small-scale datasets (Cora, Citeseer, Photo, Computers) and K-means clustering on the other datasets to produce cluster assignments for evaluation.

\begin{table*}[htbp]
  \caption{Clustering performance of \ours and 11 state-of-the-art baselines. The \textbf{boldfaced} score and \underline{underlined} score indicate the best and second-best results, respectively. ``OOM'' denotes out-of-memory, and * indicates semi-supervised method.}
  \label{table:main_clustering_resluts}
  \begin{adjustbox}{width=\textwidth,center}
    \large 
    \begin{tabular}{@{}lccccccccccccccc@{}}
      \toprule
      \multirow{2}{*}{\makecell[c]{\textbf{Method}}} & \multirow{2}{*}{\makecell[c]{\textbf{Metric}}} & \multicolumn{12}{c}{\textbf{Baselines}} & \textbf{Ours}                                                                                                                                                                                                                                                                                                                                    \\
      \cmidrule(r){3-13}
      \cmidrule(r){15-15}
                                                     &                                                & \makecell[c]{K-means}                   & \makecell[c]{Node2vec} & \makecell[c]{DGI} & \makecell[c]{GRACE}                & \makecell[c]{MVGRL}                & \makecell[c]{BGRL}                 & \makecell[c]{CCGC}                 & \makecell[c]{SCGDN}                & \makecell[c]{DGCLUSTER$^{*}$}      & \makecell[c]{S$^{3}$GC} & \makecell[c]{DMoN} &  & \makecell[c]{\ours} \\

      \midrule
      \midrule
      \addlinespace
      \multirow{4}{*}{\makecell[c]{\textbf{Cora}}}
                                                     & ACC                                            & 0.350                                   & 0.612                  & 0.726             & 0.739                              & \textbf{0.763}                     & 0.742                              & 0.664                              & 0.748                              & 0.753                              & 0.742                   & 0.517              &  & \underline{0.760}   \\
                                                     & NMI                                            & 0.173                                   & 0.444                  & 0.571             & 0.570                              & \textbf{0.608}                     & 0.584                              & 0.527                              & 0.569                              & 0.600                              & 0.588                   & 0.473              &  & \underline{0.597}   \\
                                                     & ARI                                            & 0.127                                   & 0.329                  & 0.511             & 0.527                              & \underline{0.566}                  & 0.534                              & 0.446                              & 0.526                              & 0.548                              & 0.544                   & 0.301              &  & \textbf{0.573}      \\
                                                     & F1                                             & 0.360                                   & 0.621                  & 0.692             & \underline{0.725}                  & 0.716                              & 0.691                              & 0.587                              & 0.704                              & 0.706                              & 0.721                   & 0.574              &  & \textbf{0.739}      \\
      \addlinespace
      \midrule
      \addlinespace
      \multirow{4}{*}{\makecell[c]{\textbf{Citeseer}}}
                                                     & ACC                                            & 0.421                                   & 0.421                  & 0.686             & 0.631                              & \underline{0.703}                  & 0.675                              & 0.664                              & 0.696                              & 0.618                              & 0.688                   & 0.385              &  & \textbf{0.706}      \\
                                                     & NMI                                            & 0.199                                   & 0.240                  & 0.435             & 0.399                              & \textbf{0.459}                     & 0.422                              & 0.418                              & 0.443                              & 0.373                              & 0.441                   & 0.303              &  & \underline{0.452}   \\
                                                     & ARI                                            & 0.142                                   & 0.116                  & 0.445             & 0.377                              & \textbf{0.471}                     & 0.428                              & 0.414                              & 0.454                              & 0.322                              & 0.448                   & 0.200              &  & \underline{0.468}   \\
                                                     & F1                                             & 0.394                                   & 0.401                  & 0.643             & 0.603                              & \textbf{0.654}                     & 0.631                              & 0.627                              & 0.655                              & 0.543                              & 0.643                   & 0.437              &  & \underline{0.648}   \\
      \addlinespace
      \midrule
      \addlinespace
      \multirow{4}{*}{\makecell[c]{\textbf{Photo}}}
                                                     & ACC                                            & 0.272                                   & 0.276                  & 0.430             & 0.677                              & 0.411                              & 0.665                              & 0.689                              & 0.780                              & \textbf{0.820}                     & 0.752                   & 0.248              &  & \underline{0.790}   \\
                                                     & NMI                                            & 0.132                                   & 0.115                  & 0.337             & 0.535                              & 0.303                              & 0.601                              & 0.612                              & 0.694                              & \textbf{0.735}                     & 0.598                   & 0.077              &  & \underline{0.716}   \\
                                                     & ARI                                            & 0.055                                   & 0.049                  & 0.221             & 0.427                              & 0.188                              & 0.441                              & 0.495                              & 0.607                              & \textbf{0.671}                     & 0.561                   & 0.038              &  & \underline{0.615}   \\
                                                     & F1                                             & 0.240                                   & 0.215                  & 0.352             & 0.503                              & 0.329                              & 0.631                              & 0.609                              & 0.716                              & \textbf{0.752}                     & 0.729                   & 0.180              &  & \underline{0.729}   \\
      \addlinespace
      \midrule
      \addlinespace
      \multirow{4}{*}{\makecell[c]{\textbf{Computers}}}
                                                     & ACC                                            & 0.225                                   & 0.356                  & 0.479             & 0.519                              & 0.580                              & 0.469                              & 0.480                              & 0.582                              & 0.453                              & \underline{0.588}       & 0.432              &  & \textbf{0.620}      \\
                                                     & NMI                                            & 0.110                                   & 0.278                  & 0.420             & 0.538                              & 0.482                              & 0.441                              & 0.452                              & 0.545                              & 0.498                              & \underline{0.560}       & 0.461              &  & \textbf{0.592}      \\
                                                     & ARI                                            & 0.056                                   & 0.248                  & 0.306             & 0.343                              & 0.433                              & 0.306                              & 0.290                              & 0.430                              & 0.261                              & \underline{0.438}       & 0.288              &  & \textbf{0.462}      \\
                                                     & F1                                             & 0.152                                   & 0.224                  & 0.390             & 0.390                              & 0.405                              & 0.415                              & 0.380                              & 0.480                              & 0.372                              & \underline{0.475}       & 0.390              &  & \textbf{0.574}      \\
      \addlinespace
      \midrule
      \addlinespace
      \multirow{4}{*}{\makecell[c]{\textbf{ogbn-arxiv}}}
                                                     & ACC                                            & 0.181                                   & 0.290                  & 0.314             & \multirow{4}{*}{\makecell[c]{OOM}} & \multirow{4}{*}{\makecell[c]{OOM}} & \multirow{4}{*}{\makecell[c]{OOM}} & \multirow{4}{*}{\makecell[c]{OOM}} & \multirow{4}{*}{\makecell[c]{OOM}} & \multirow{4}{*}{\makecell[c]{OOM}} & \underline{0.350}       & 0.250              &  & \textbf{0.388}      \\
                                                     & NMI                                            & 0.221                                   & 0.406                  & 0.412             &                                    &                                    &                                    &                                    &                                    &                                    & \underline{0.463}       & 0.356              &  & \textbf{0.469}      \\
                                                     & ARI                                            & 0.074                                   & 0.190                  & 0.223             &                                    &                                    &                                    &                                    &                                    &                                    & \underline{0.270}       & 0.127              &  & \textbf{0.310}      \\
                                                     & F1                                             & 0.129                                   & 0.220                  & \underline{0.230} &                                    &                                    &                                    &                                    &                                    &                                    & \underline{0.230}       & 0.190              &  & \textbf{0.266}      \\
      \addlinespace
      \midrule
      \addlinespace
      \multirow{4}{*}{\makecell[c]{\textbf{Reddit}}}
                                                     & ACC                                            & 0.089                                   & 0.709                  & 0.224             & \multirow{4}{*}{\makecell[c]{OOM}} & \multirow{4}{*}{\makecell[c]{OOM}} & \multirow{4}{*}{\makecell[c]{OOM}} & \multirow{4}{*}{\makecell[c]{OOM}} & \multirow{4}{*}{\makecell[c]{OOM}} & \multirow{4}{*}{\makecell[c]{OOM}} & \underline{0.736}       & 0.529              &  & \textbf{0.911}      \\
                                                     & NMI                                            & 0.114                                   & 0.792                  & 0.306             &                                    &                                    &                                    &                                    &                                    &                                    & \underline{0.807}       & 0.628              &  & \textbf{0.875}      \\
                                                     & ARI                                            & 0.029                                   & 0.640                  & 0.170             &                                    &                                    &                                    &                                    &                                    &                                    & \underline{0.745}       & 0.502              &  & \textbf{0.907}      \\
                                                     & F1                                             & 0.068                                   & 0.551                  & 0.183             &                                    &                                    &                                    &                                    &                                    &                                    & \underline{0.560}       & 0.260              &  & \textbf{0.853}      \\
      \addlinespace
      \midrule
      \addlinespace
      \multirow{4}{*}{\makecell[c]{\textbf{ogbn-products}}}
                                                     & ACC                                            & 0.200                                   & 0.357                  & 0.320             & \multirow{4}{*}{\makecell[c]{OOM}} & \multirow{4}{*}{\makecell[c]{OOM}} & \multirow{4}{*}{\makecell[c]{OOM}} & \multirow{4}{*}{\makecell[c]{OOM}} & \multirow{4}{*}{\makecell[c]{OOM}} & \multirow{4}{*}{\makecell[c]{OOM}} & \underline{0.402}       & 0.304              &  & \textbf{0.425}      \\
                                                     & NMI                                            & 0.273                                   & 0.489                  & 0.467             &                                    &                                    &                                    &                                    &                                    &                                    & \underline{0.536}       & 0.428              &  & \textbf{0.551}      \\
                                                     & ARI                                            & 0.082                                   & 0.170                  & 0.174             &                                    &                                    &                                    &                                    &                                    &                                    & \textbf{0.230}          & 0.139              &  & \underline{0.215}   \\
                                                     & F1                                             & 0.124                                   & 0.247                  & 0.192             &                                    &                                    &                                    &                                    &                                    &                                    & \underline{0.250}       & 0.210              &  & \textbf{0.276}      \\
      \addlinespace
      \bottomrule
    \end{tabular}
  \end{adjustbox}
\end{table*}

\subsection{Experimental setup}
We use a single NVIDIA A100 GPU with 40GB memory for each method. All experiments are repeated 5 times and the mean values are reported in Table~\ref{table:main_clustering_resluts}. For all datasets except for ogbn-papers100M, we set the number of run epochs at 400 and limit the maximum runtime to 1 hour. For ogbn-papers100M we allow up to 6 hours of training in addition to 256GB memory limitation.
We employ full batch training on small-scale graphs and set
the number of rooted nodes $n$ as 2048 for large and extra-large datasets. Due to space limitations, more details about the hyperparameters are mentioned in Table~\ref{tab:hyperparameter} in the Appendix. We have provided a mini-batch and highly scalable implementation of \ours in PyTorch, making it easy for experiments with all datasets to adapt to GPU processing. For all datasets, it is only need to store the subgraphs and corresponding modularity matrix in a sparse form to execute \ours's \textit{forward} and \textit{backward} pass on the GPU. We also provide the GPU memory cost and the time required to execute the training process for each method in Table~\ref{table:complexity} in Appendix.

\section{Experimental results}
In this section, we present the experimental results of \ours in the graph clustering task. In addition to the main experiments, we have conducted supplementary experiments aimed at answering the following research questions:

\textbf{(Q1):} Can \ours scale to extra-large scale graphs containing up to 100 million nodes and outperform state-of-the-art baselines?

\textbf{(Q2):} Can modularity maximization pretext tasks effectively mitigate the problem of semantic drift?

\textbf{(Q3):} Is \ours robust to different hyperparameters, and does each mechanism within \ours contribute uniquely to its performance?

\begin{table}[!h]
  \centering
  \caption{Comparison results of different scalable methods on ogbn-papers100M with 111M nodes and 1.6B edges.}
  \label{tab:ogbn_papers100m}
  \begin{tabular}{lcccc}
    \toprule
    \multirow{2}{*}{}  & \multicolumn{4}{c}{ogbn-papers100M}                                                             \\
    \cmidrule{2-5}
                       & Accuracy                            & NMI               & ARI               & F1                \\
    \midrule
    \midrule
    \textbf{K-means}   & 0.144                               & 0.368             & 0.074             & \underline{0.101} \\
    \textbf{Node2vec}  & \underline{0.175}                   & 0.380             & \underline{0.112} & 0.099             \\
    \textbf{DGI}       & 0.151                               & 0.416             & 0.096             & 0.111             \\
    \textbf{S$^{3}$GC} & 0.173                               & \underline{0.453} & 0.110             & \textbf{0.118}    \\
    \textbf{\ours}     & \textbf{0.288}                      & \textbf{0.463}    & \textbf{0.207}    & 0.096             \\
    \bottomrule
  \end{tabular}
\end{table}

\begin{figure}[t]
  \centering
  \includegraphics[width=\linewidth]{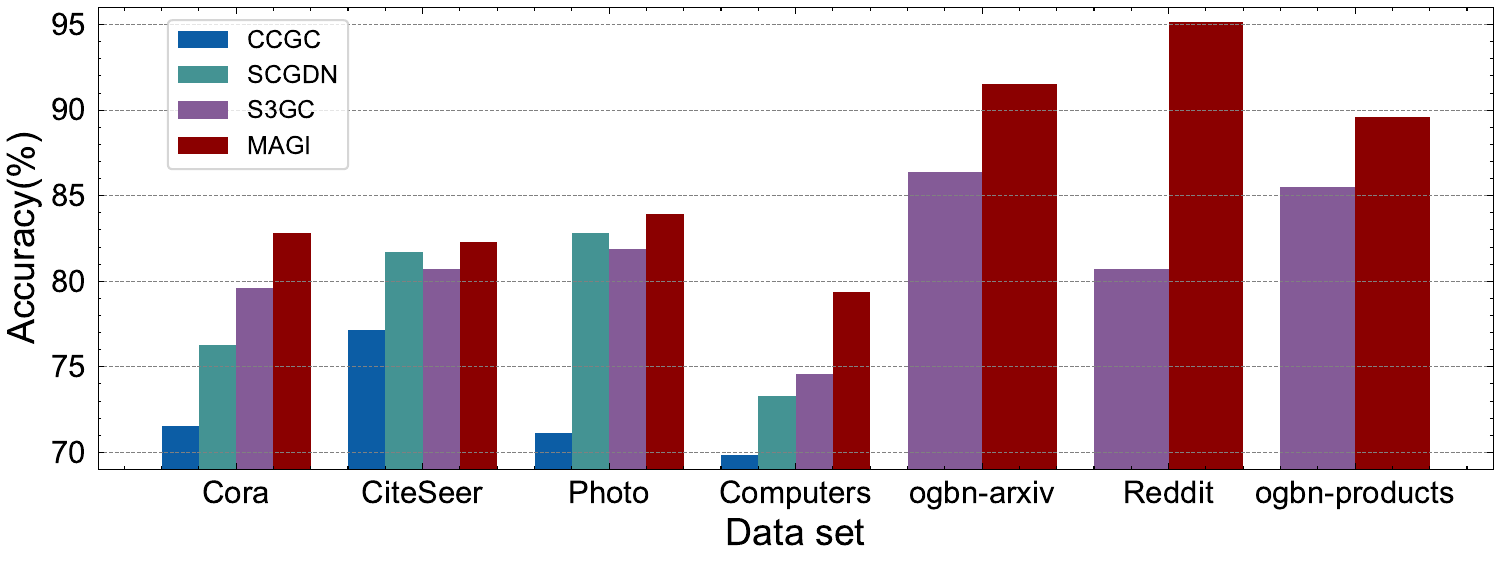}
  \caption{Comparison of pseudo-labels constructed by \ours and other methods.}
  \label{fig:semantic_drift}
\end{figure}

\begin{figure}[t]
  \centering
  \begin{subfigure}{.5\columnwidth}
    \centering
    \includegraphics[width=\linewidth]{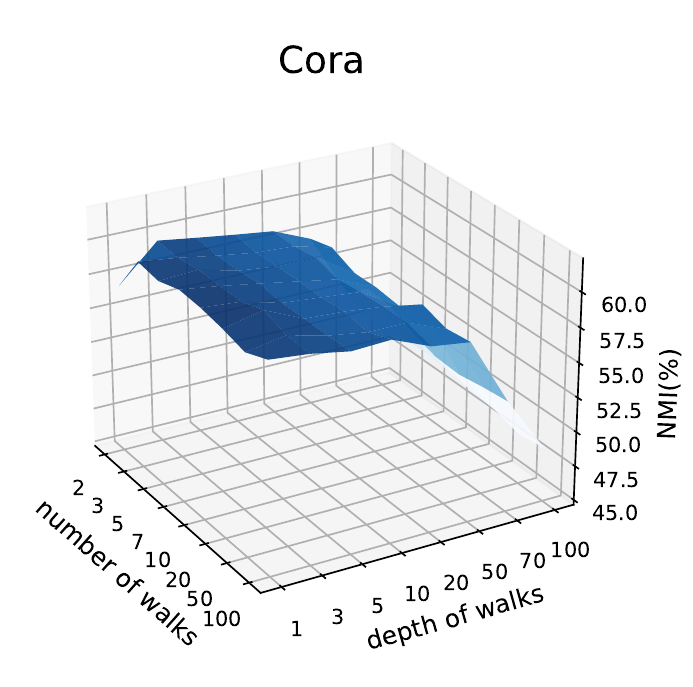}
    \label{fig:ablation_randomwalk_cora}
  \end{subfigure}%
  \begin{subfigure}{.5\columnwidth}
    \centering
    \includegraphics[width=\linewidth]{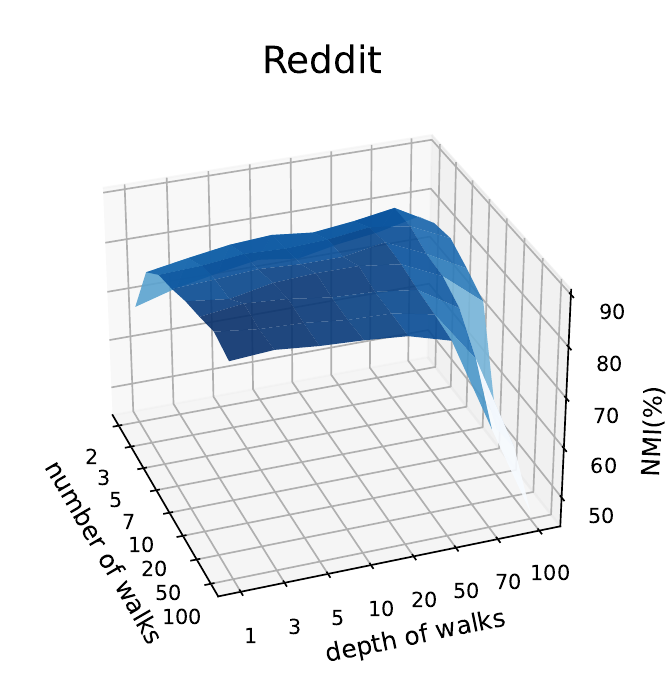}
    \label{fig:ablation_randomwalk_reddit}
  \end{subfigure}
  \caption{The performance of \ours with varying the number and depth of random walks on the Cora and Reddit dataset in terms of NMI.}
  \label{fig:ablation_randomwalk}
\end{figure}

\subsection{Graph clustering result}
Table~\ref{table:main_clustering_resluts} presents a comparison of \ours's clustering performance against various baseline methods across different scale datasets. For small-scale datasets, namely Cora, Citeseer, Photo, and Computers, we observe that MVGRL, S$^{3}$GC, and DGCLUSTER emerge as the three strongest baseline methods.
Nonetheless, we find that \ours consistently ranks first or second in performance in most cases. We note that even when compared to semi-supervised methods such as DGCLUSTER, \ours is only slightly behind on the Photo dataset and maintains a consistent lead on the other three datasets, demonstrating \ours’s superior performance.
Next, we observe the performance on large-scale datasets and find that \ours notably outperforms baseline methods like DGI, DMoN and S$^{3}$GC.
\ours is $\thicksim3.6\%$ better on ogbn-arxiv, $\thicksim29\%$ better on Reddit and $\thicksim2.6\%$ better on ogbn-products in terms of clustering F1 as compared to the second-best method.
Significantly, \ours outperforms the second-best method on the Reddit dataset with an approximate increase of $\thicksim18\%$ in Accuracy, $\thicksim7\%$ in NMI, $\thicksim16\%$ in ARI, and $\thicksim29\%$ in F1 score. One possible reason for the superior performance is that the Reddit dataset exhibits a higher graph density, which results in a more distinct community structure.

\textbf{ogbn-papers100M:} To answer \textbf{Q1}, we conduct a comparative analysis of \ours' performance on an extra-large dataset containing 111M nodes and 1.6B edges. Note that we compare K-means, Node2vec, DGI, and S$^{3}$GC since others can not scale to this dataset. The results are presented in Table~\ref{tab:ogbn_papers100m}. Our experimental results reveal that \ours adeptly scales to this dataset and demonstrates a significant performance improvement over methods relying solely on features (K-means) by approximately $\thicksim14.4\%$, exclusively on graph structure (Node2vec) by approximately $\thicksim11.3\%$, and both features and structure (S$^{3}$GC) by approximately $\thicksim11.5\%$ in terms of accuracy on the ogbn-papers100M dataset.

\subsection{Semantic drift mitigation}\label{SemanticDriftExperiment}
In practice, it is difficult to measure the degree of semantic drift in a graph. To answer \textbf{Q2}, we alternatively measure the semantic drift based on the quality of pseudo-labels generated by various methods. Specifically, considering the augmentation-free GCLs referenced in Section~\ref{Baseline}, we employ ground-truth labels to evaluate the accuracy of pseudo-labels produced by these methods. In this context, a positive sample pair is considered as `1' if both nodes belong to the same class, while a negative sample pair is deemed `1' if the nodes belong to different classes; otherwise, they are labeled as `0'. The results (refer to Figure~\ref{fig:semantic_drift}) demonstrate that, in comparison to other augmentation-free GCLs, \ours generates higher-quality pseudo-labels. This confirms the efficacy of modularity maximization as a pretext task in mitigating semantic drift.

\begin{figure}[t]
  \centering
  \begin{subfigure}{.5\columnwidth}
    \centering
    \includegraphics[width=\linewidth]{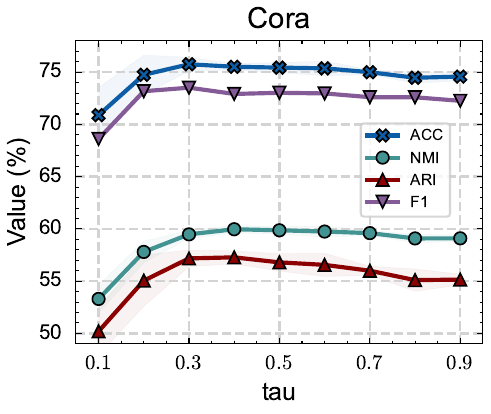}
    \label{fig:ablation_tau_cora}
  \end{subfigure}%
  \begin{subfigure}{.5\columnwidth}
    \centering
    \includegraphics[width=\linewidth]{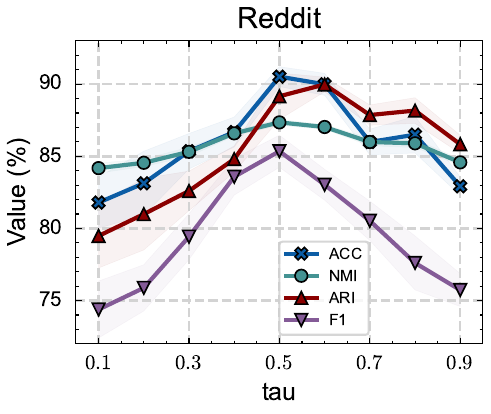}
    \label{fig:ablation_tau_reddit}
  \end{subfigure}
  \caption{The performance of \ours with varying the temperature $\tau$ on the Cora and Reddit dataset, respectively.}
  \label{fig:ablation_tau}
\end{figure}

\begin{figure}[t]
  \centering
  \includegraphics[width=\linewidth]{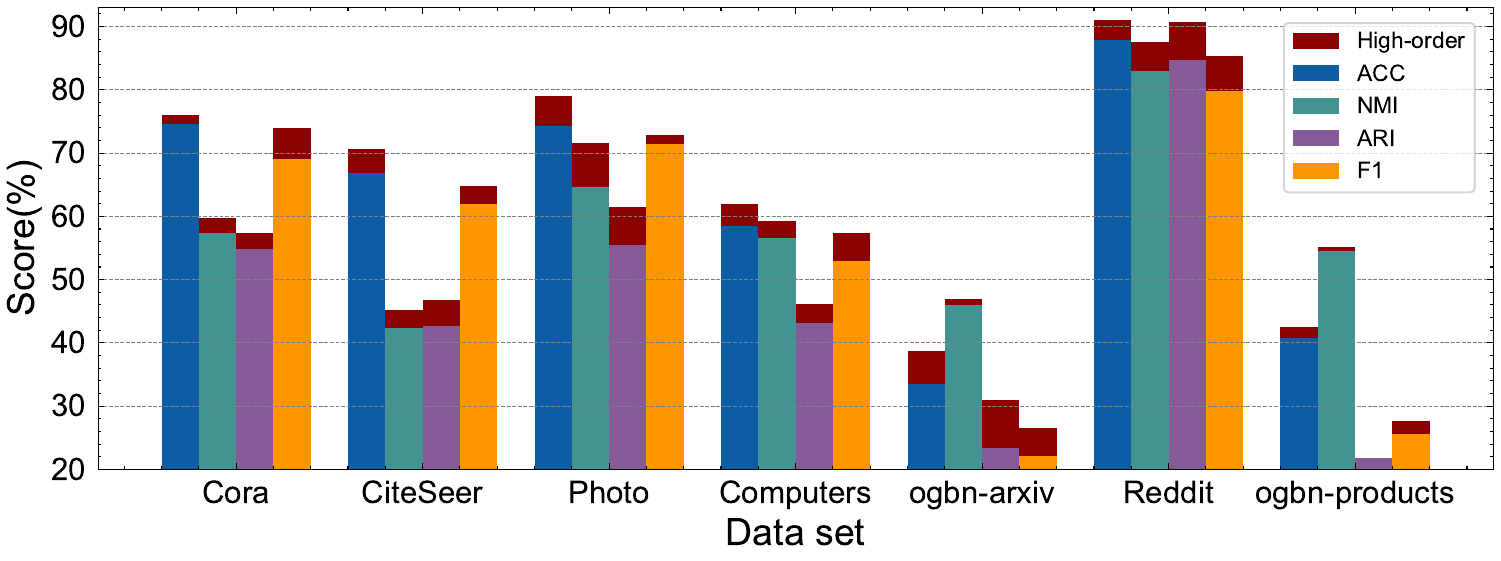}
  \caption{Benefits of incorporating high-order proximity, with the red part of each bar representing the gains provided by high-order proximity.}
  \label{fig:high_order}
\end{figure}

\begin{figure*}[t]
  \centering
  \includegraphics[width=1\textwidth]{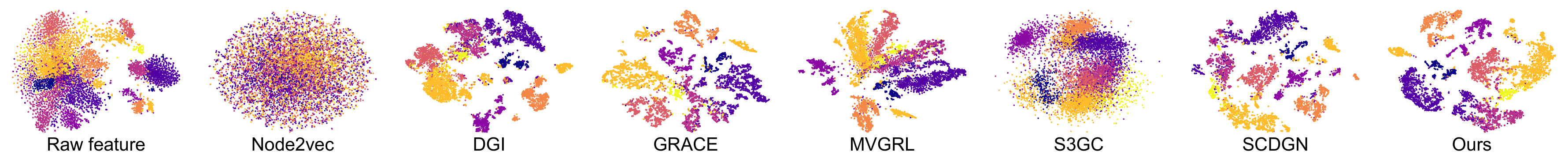}
  \caption{$t$-SNE visualization of eight unsupervised methods on the Photo dataset.}
  \label{Fig_visualization}
\end{figure*}

\begin{figure}[t]
  \centering
  \includegraphics[width=\linewidth]{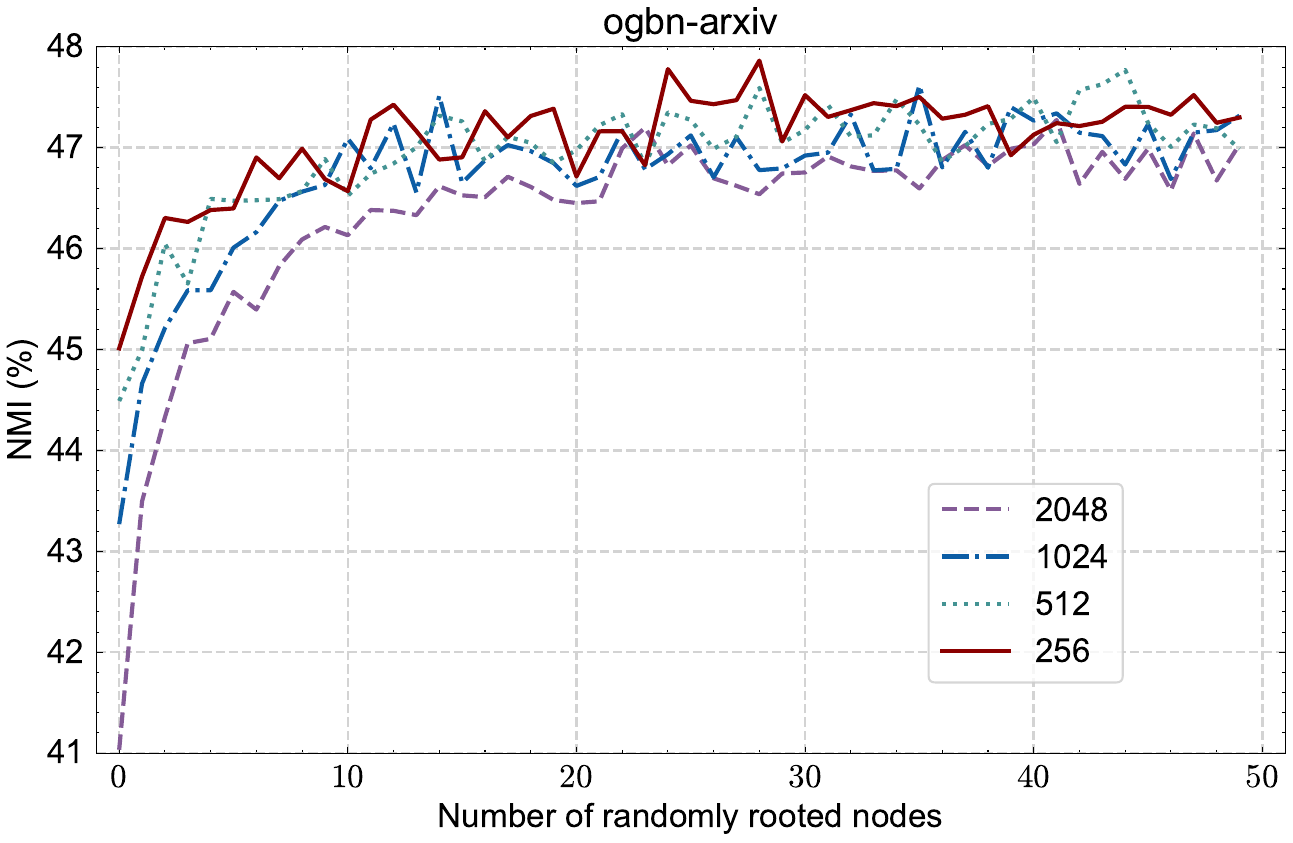}
  \caption{The performance of \ours with varying different number of randomly rooted nodes $n$ on the ogbn-arxiv dataset in terms of NMI.}
  \label{fig:ablation_batchsize}
\end{figure}

\subsection{Ablation study}
To answer \textbf{Q3}, we first conduct thorough ablation studies on hyperparameters including the number of walks $t$, depth of walks $l$, number of randomly rooted nodes $n$ and temperature $\tau$ to examine the stability of \ours’s clustering.
Our findings are listed below:

(1) The performance of \ours remains relatively stable as long as the number and depth of random walks are not excessively extreme, as shown in Figure~\ref{fig:ablation_randomwalk}. We also note that the walk depth $\thicksim5$ is an optimal choice, as a larger walk depth tends to sample positive pairs with differing community semantics. This finding is consistent with the previous work~\cite{devvrit2022s3gc}.

(2) \ours can achieve the best performance when the temperature $\tau$ is around $\thicksim0.5$ (see Figure~\ref{fig:ablation_tau}). A temperature $\tau$ that is too small or too large can lead to the model paying too much or too little attention to hard negative samples~\cite{Wang2021Understanding}, respectively.

(3) Different number of randomly sampled root nodes $n$ have similar clustering performance, but a smaller $n$ can lead to faster convergence (see Figure~\ref{fig:ablation_batchsize}). Combined with \ours's stable performance at smaller walk depths, this demonstrates \ours's good scalability, meaning that for each node, \ours only needs to sample a few positive samples.

\textbf{High-order proximity.}
We then showcase the effectiveness of exploring high-order proximity in \ours.
In a given batch $\mathcal{B}$, the effectiveness of employing random walks versus directly sampling a submatrix from the vanilla modularity matrix is compared across various datasets. Figure~\ref{fig:high_order} demonstrates that, in the majority of cases, the use of random walks outperforms direct sampling. This result highlights the advantage of incorporating high-order proximity into the analysis, enabling a better capture of the complex relationships and community structure within the graph.

\textbf{Effectiveness of different components.}
We conduct ablation study to manifest the efficacy of different components in \ours. We set five variants of our model for comparison. Results are shown in Table~\ref{tab:abltion} in Appendix, where ``\ours(w/ SL)'' refers to the use of simple contrastive loss defined in Eq.(~\ref{eq:cl_loss_simple}),  ``\ours(w/ MS)'' refers to use \textbf{S1} and the sign of modularity score to define positive and negative sample sets, ``\ours(w/ EI)'' refers to use \textbf{S1} and edge indicators to define positive and negative sample sets, ``\ours(w/ HMS)'' refers to use sign of high-order modularity score~\cite{Liu2022Robust} to define positive and negative sample sets and \ours refers to use \textbf{S1} and \textbf{S2} to define positive and negative sample sets.

In Table~\ref{tab:abltion}, it is observed that each improvement of our model contributes to the final performance. First, loss~(\ref{eq:SimCLR_loss}) performs better than loss~(\ref{eq:cl_loss_simple}). Secondly, the direct use of edge indicators to define the positive and negative sample sets achieves the worst effect. We infer that this may be because some edges exist between different communities, which leads to some positive sample pairs from different communities, increasing the impact of semantic drift. This can be mitigated by the use of modularity scores, and consideration of high-order proximity in the community can further eliminate the impact of semantic drift. Finally, our sampling strategy achieves similar performance compared to using the sign of high order modularity matrix proposed in~\cite{Liu2022Robust}, but our sampling strategy has lower computational complexity and can be scaled to large-scale graph datasets with 100M nodes.

\subsection{Visualization}

In this part, we measure the quality of the generated embeddings by directly employing t-SNE~\cite{tsne}. The generated embeddings of each method are projected into 2-dimensional vectors for visualization.
Due to space limitations, we have selected seven strong baseline methods for visual analysis. It should also be noted that DGCLUSTER~\cite{DGCLUSTER2024AAAI}, being a semi-supervised method, was excluded from this comparison.
The visualization clearly demonstrates that \ours produces representations with significantly higher clustering efficacy compared to baseline methods.

\section{CONCLUSION}
In this paper, we explore the problem of graph clustering via neural modularity maximization.
Our work establishes the connection between neural modularity maximization and graph contrastive learning. This insight motivates us to propose \ours, a community-aware graph clustering framework with modularity maximization as the pretext task for contrastive learning.
\ours is an augmentation-free GCL framework, which avoids potential semantic drift and scalability issues.
To ensure better scalability, \ours adopts a two-stage random walk to approximate the modularity matrix in a mini-batch manner, followed by a principled contrastive loss to optimize the goal of modularity maximization.
Extensive experiments on eight real-world graph datasets demonstrate the effectiveness of our method, which achieves state-of-the-art in most cases compared with strong baselines.
We hope that the straightforward nature of our approach serves as a reminder to the community to reevaluate simpler alternatives that may have been overlooked, thereby inspiring future research.

\clearpage

\bibliographystyle{ACM-Reference-Format}
\bibliography{main}


\begin{thebibliography}{51}


\ifx \showCODEN    \undefined \def \showCODEN     #1{\unskip}     \fi
\ifx \showDOI      \undefined \def \showDOI       #1{#1}\fi
\ifx \showISBNx    \undefined \def \showISBNx     #1{\unskip}     \fi
\ifx \showISBNxiii \undefined \def \showISBNxiii  #1{\unskip}     \fi
\ifx \showISSN     \undefined \def \showISSN      #1{\unskip}     \fi
\ifx \showLCCN     \undefined \def \showLCCN      #1{\unskip}     \fi
\ifx \shownote     \undefined \def \shownote      #1{#1}          \fi
\ifx \showarticletitle \undefined \def \showarticletitle #1{#1}   \fi
\ifx \showURL      \undefined \def \showURL       {\relax}        \fi
\providecommand\bibfield[2]{#2}
\providecommand\bibinfo[2]{#2}
\providecommand\natexlab[1]{#1}
\providecommand\showeprint[2][]{arXiv:#2}

\bibitem[Bhatia and Rani(2018)]%
        {Bhatia2018DFuzzyAD}
\bibfield{author}{\bibinfo{person}{Vandana Bhatia} {and} \bibinfo{person}{Rinkle Rani}.} \bibinfo{year}{2018}\natexlab{}.
\newblock \showarticletitle{DFuzzy: a deep learning-based fuzzy clustering model for large graphs}.
\newblock \bibinfo{journal}{\emph{Knowledge and Information Systems}}  \bibinfo{volume}{57} (\bibinfo{year}{2018}), \bibinfo{pages}{159--181}.
\newblock


\bibitem[Bhowmick et~al\mbox{.}(2024)]%
        {DGCLUSTER2024AAAI}
\bibfield{author}{\bibinfo{person}{Aritra Bhowmick}, \bibinfo{person}{Mert Kosan}, \bibinfo{person}{Zexi Huang}, \bibinfo{person}{Ambuj~K. Singh}, {and} \bibinfo{person}{Sourav Medya}.} \bibinfo{year}{2024}\natexlab{}.
\newblock \showarticletitle{{DGCLUSTER:} {A} Neural Framework for Attributed Graph Clustering via Modularity Maximization}. In \bibinfo{booktitle}{\emph{{AAAI}}}. \bibinfo{publisher}{{AAAI} Press}, \bibinfo{pages}{11069--11077}.
\newblock


\bibitem[Blondel et~al\mbox{.}(2008)]%
        {louvain}
\bibfield{author}{\bibinfo{person}{Vincent~D Blondel}, \bibinfo{person}{Jean-Loup Guillaume}, \bibinfo{person}{Renaud Lambiotte}, {and} \bibinfo{person}{Etienne Lefebvre}.} \bibinfo{year}{2008}\natexlab{}.
\newblock \showarticletitle{Fast unfolding of communities in large networks}.
\newblock \bibinfo{journal}{\emph{Journal of statistical mechanics: theory and experiment}} \bibinfo{volume}{2008}, \bibinfo{number}{10} (\bibinfo{year}{2008}), \bibinfo{pages}{P10008}.
\newblock


\bibitem[Brandes et~al\mbox{.}(2006)]%
        {Brandes2006MaximizingMI}
\bibfield{author}{\bibinfo{person}{Ulrik Brandes}, \bibinfo{person}{Daniel Delling}, \bibinfo{person}{Marco Gaertler}, \bibinfo{person}{Rachelle Goerke}, \bibinfo{person}{Martin Hoefer}, \bibinfo{person}{Zoran Nikoloski}, {and} \bibinfo{person}{Donald Wagner}.} \bibinfo{year}{2006}\natexlab{}.
\newblock \showarticletitle{Maximizing Modularity is hard}.
\newblock \bibinfo{journal}{\emph{arXiv: Data Analysis, Statistics and Probability}} (\bibinfo{year}{2006}).
\newblock


\bibitem[Brede(2012)]%
        {Brede2012NetworksAnIM}
\bibfield{author}{\bibinfo{person}{Markus Brede}.} \bibinfo{year}{2012}\natexlab{}.
\newblock \showarticletitle{Networks—An Introduction. Mark E. J. Newman. (2010, Oxford University Press.) \$65.38, £35.96 (hardcover), 772 pages. ISBN-978-0-19-920665-0.}
\newblock \bibinfo{journal}{\emph{Artificial Life}}  \bibinfo{volume}{18} (\bibinfo{year}{2012}), \bibinfo{pages}{241--242}.
\newblock


\bibitem[Chen et~al\mbox{.}(2020)]%
        {simclr}
\bibfield{author}{\bibinfo{person}{Ting Chen}, \bibinfo{person}{Simon Kornblith}, \bibinfo{person}{Mohammad Norouzi}, {and} \bibinfo{person}{Geoffrey Hinton}.} \bibinfo{year}{2020}\natexlab{}.
\newblock \showarticletitle{A simple framework for contrastive learning of visual representations}. In \bibinfo{booktitle}{\emph{ICML}}. Article \bibinfo{articleno}{149}, \bibinfo{numpages}{11}~pages.
\newblock


\bibitem[Choong et~al\mbox{.}(2018)]%
        {Choong2018Learning}
\bibfield{author}{\bibinfo{person}{Jun~Jin Choong}, \bibinfo{person}{Xin Liu}, {and} \bibinfo{person}{Tsuyoshi Murata}.} \bibinfo{year}{2018}\natexlab{}.
\newblock \showarticletitle{Learning Community Structure with Variational Autoencoder}. In \bibinfo{booktitle}{\emph{ICDM}}. \bibinfo{pages}{69--78}.
\newblock


\bibitem[Cui et~al\mbox{.}(2020)]%
        {AGE2020KDD}
\bibfield{author}{\bibinfo{person}{Ganqu Cui}, \bibinfo{person}{Jie Zhou}, \bibinfo{person}{Cheng Yang}, {and} \bibinfo{person}{Zhiyuan Liu}.} \bibinfo{year}{2020}\natexlab{}.
\newblock \showarticletitle{Adaptive Graph Encoder for Attributed Graph Embedding}. In \bibinfo{booktitle}{\emph{KDD}}. \bibinfo{pages}{976–985}.
\newblock


\bibitem[Devvrit et~al\mbox{.}(2022)]%
        {devvrit2022s3gc}
\bibfield{author}{\bibinfo{person}{Fnu Devvrit}, \bibinfo{person}{Aditya Sinha}, \bibinfo{person}{Inderjit~S Dhillon}, {and} \bibinfo{person}{Prateek Jain}.} \bibinfo{year}{2022}\natexlab{}.
\newblock \showarticletitle{S3{GC}: Scalable Self-Supervised Graph Clustering}. In \bibinfo{booktitle}{\emph{NeurIPS}}, \bibfield{editor}{\bibinfo{person}{Alice~H. Oh}, \bibinfo{person}{Alekh Agarwal}, \bibinfo{person}{Danielle Belgrave}, {and} \bibinfo{person}{Kyunghyun Cho}} (Eds.).
\newblock


\bibitem[Felzenszwalb and Huttenlocher(2004)]%
        {Felzenszwalb2004EfficientGI}
\bibfield{author}{\bibinfo{person}{Pedro~F. Felzenszwalb} {and} \bibinfo{person}{Daniel~P. Huttenlocher}.} \bibinfo{year}{2004}\natexlab{}.
\newblock \showarticletitle{Efficient Graph-Based Image Segmentation}.
\newblock \bibinfo{journal}{\emph{International Journal of Computer Vision}}  \bibinfo{volume}{59} (\bibinfo{year}{2004}), \bibinfo{pages}{167--181}.
\newblock


\bibitem[Grill et~al\mbox{.}(2020)]%
        {BYOL}
\bibfield{author}{\bibinfo{person}{Jean-Bastien Grill}, \bibinfo{person}{Florian Strub}, \bibinfo{person}{Florent Altch\'{e}}, \bibinfo{person}{Corentin Tallec}, \bibinfo{person}{Pierre~H. Richemond}, \bibinfo{person}{Elena Buchatskaya}, \bibinfo{person}{Carl Doersch}, \bibinfo{person}{Bernardo~Avila Pires}, \bibinfo{person}{Zhaohan~Daniel Guo}, \bibinfo{person}{Mohammad~Gheshlaghi Azar}, \bibinfo{person}{Bilal Piot}, \bibinfo{person}{Koray Kavukcuoglu}, \bibinfo{person}{R\'{e}mi Munos}, {and} \bibinfo{person}{Michal Valko}.} \bibinfo{year}{2020}\natexlab{}.
\newblock \showarticletitle{Bootstrap your own latent a new approach to self-supervised learning}. In \bibinfo{booktitle}{\emph{NeurIPS}}. \bibinfo{numpages}{14}~pages.
\newblock


\bibitem[Grover and Leskovec(2016)]%
        {node2vec}
\bibfield{author}{\bibinfo{person}{Aditya Grover} {and} \bibinfo{person}{Jure Leskovec}.} \bibinfo{year}{2016}\natexlab{}.
\newblock \showarticletitle{node2vec: Scalable Feature Learning for Networks}. In \bibinfo{booktitle}{\emph{KDD}}. \bibinfo{publisher}{Association for Computing Machinery}, \bibinfo{pages}{855–864}.
\newblock


\bibitem[Hadsell et~al\mbox{.}(2006)]%
        {Hadsell2006Dimensionality}
\bibfield{author}{\bibinfo{person}{R. Hadsell}, \bibinfo{person}{S. Chopra}, {and} \bibinfo{person}{Y. LeCun}.} \bibinfo{year}{2006}\natexlab{}.
\newblock \showarticletitle{Dimensionality Reduction by Learning an Invariant Mapping}. In \bibinfo{booktitle}{\emph{2006 IEEE Computer Society Conference on Computer Vision and Pattern Recognition (CVPR'06)}}, Vol.~\bibinfo{volume}{2}. \bibinfo{pages}{1735--1742}.
\newblock


\bibitem[Hamilton et~al\mbox{.}(2017)]%
        {GraphSAGE}
\bibfield{author}{\bibinfo{person}{William~L. Hamilton}, \bibinfo{person}{Rex Ying}, {and} \bibinfo{person}{Jure Leskovec}.} \bibinfo{year}{2017}\natexlab{}.
\newblock \showarticletitle{Inductive Representation Learning on Large Graphs}. In \bibinfo{booktitle}{\emph{NeurIPS}}. \bibinfo{pages}{1025–1035}.
\newblock


\bibitem[Hassani and Khasahmadi(2020)]%
        {MVGRL2020ICML}
\bibfield{author}{\bibinfo{person}{Kaveh Hassani} {and} \bibinfo{person}{Amir~Hosein Khasahmadi}.} \bibinfo{year}{2020}\natexlab{}.
\newblock \showarticletitle{Contrastive multi-view representation learning on graphs}. In \bibinfo{booktitle}{\emph{ICML}} \emph{(\bibinfo{series}{ICML'20})}. \bibinfo{publisher}{JMLR.org}, Article \bibinfo{articleno}{385}, \bibinfo{numpages}{11}~pages.
\newblock


\bibitem[Hjelm et~al\mbox{.}(2019)]%
        {hjelm2018learning}
\bibfield{author}{\bibinfo{person}{R~Devon Hjelm}, \bibinfo{person}{Alex Fedorov}, \bibinfo{person}{Samuel Lavoie-Marchildon}, \bibinfo{person}{Karan Grewal}, \bibinfo{person}{Phil Bachman}, \bibinfo{person}{Adam Trischler}, {and} \bibinfo{person}{Yoshua Bengio}.} \bibinfo{year}{2019}\natexlab{}.
\newblock \showarticletitle{Learning deep representations by mutual information estimation and maximization}. In \bibinfo{booktitle}{\emph{ICLR}}.
\newblock


\bibitem[Hu et~al\mbox{.}(2020)]%
        {hu2020ogb}
\bibfield{author}{\bibinfo{person}{Weihua Hu}, \bibinfo{person}{Matthias Fey}, \bibinfo{person}{Marinka Zitnik}, \bibinfo{person}{Yuxiao Dong}, \bibinfo{person}{Hongyu Ren}, \bibinfo{person}{Bowen Liu}, \bibinfo{person}{Michele Catasta}, {and} \bibinfo{person}{Jure Leskovec}.} \bibinfo{year}{2020}\natexlab{}.
\newblock \showarticletitle{Open Graph Benchmark: Datasets for Machine Learning on Graphs}. In \bibinfo{booktitle}{\emph{NeurIPS}}.
\newblock


\bibitem[Kipf and Welling(2016)]%
        {VGAE}
\bibfield{author}{\bibinfo{person}{Thomas~N Kipf} {and} \bibinfo{person}{Max Welling}.} \bibinfo{year}{2016}\natexlab{}.
\newblock \showarticletitle{Variational Graph Auto-Encoders}.
\newblock \bibinfo{journal}{\emph{NeurIPS Workshop on Bayesian Deep Learning}} (\bibinfo{year}{2016}).
\newblock


\bibitem[Kipf and Welling(2017)]%
        {GCN}
\bibfield{author}{\bibinfo{person}{Thomas~N. Kipf} {and} \bibinfo{person}{Max Welling}.} \bibinfo{year}{2017}\natexlab{}.
\newblock \showarticletitle{Semi-Supervised Classification with Graph Convolutional Networks}. In \bibinfo{booktitle}{\emph{ICLR}}.
\newblock


\bibitem[Lee et~al\mbox{.}(2022)]%
        {AFGRL}
\bibfield{author}{\bibinfo{person}{Namkyeong Lee}, \bibinfo{person}{Junseok Lee}, {and} \bibinfo{person}{Chanyoung Park}.} \bibinfo{year}{2022}\natexlab{}.
\newblock \showarticletitle{Augmentation-Free Self-Supervised Learning on Graphs}.
\newblock  (\bibinfo{year}{2022}).
\newblock


\bibitem[Li et~al\mbox{.}(2023)]%
        {maskgae}
\bibfield{author}{\bibinfo{person}{Jintang Li}, \bibinfo{person}{Ruofan Wu}, \bibinfo{person}{Wangbin Sun}, \bibinfo{person}{Liang Chen}, \bibinfo{person}{Sheng Tian}, \bibinfo{person}{Liang Zhu}, \bibinfo{person}{Changhua Meng}, \bibinfo{person}{Zibin Zheng}, {and} \bibinfo{person}{Weiqiang Wang}.} \bibinfo{year}{2023}\natexlab{}.
\newblock \showarticletitle{What's Behind the Mask: Understanding Masked Graph Modeling for Graph Autoencoders}. In \bibinfo{booktitle}{\emph{{KDD}}}. \bibinfo{publisher}{{ACM}}, \bibinfo{pages}{1268--1279}.
\newblock


\bibitem[Li et~al\mbox{.}(2024)]%
        {spikegcl}
\bibfield{author}{\bibinfo{person}{Jintang Li}, \bibinfo{person}{Huizhe Zhang}, \bibinfo{person}{Ruofan Wu}, \bibinfo{person}{Zulun Zhu}, \bibinfo{person}{Baokun Wang}, \bibinfo{person}{Changhua Meng}, \bibinfo{person}{Zibin Zheng}, {and} \bibinfo{person}{Liang Chen}.} \bibinfo{year}{2024}\natexlab{}.
\newblock \showarticletitle{A Graph is Worth 1-bit Spikes: When Graph Contrastive Learning Meets Spiking Neural Networks}. In \bibinfo{booktitle}{\emph{ICLR}}.
\newblock


\bibitem[Liu et~al\mbox{.}(2022)]%
        {Liu2022Robust}
\bibfield{author}{\bibinfo{person}{Yunfei Liu}, \bibinfo{person}{Zhen Liu}, \bibinfo{person}{Xiaodong Feng}, {and} \bibinfo{person}{Zhongyi Li}.} \bibinfo{year}{2022}\natexlab{}.
\newblock \showarticletitle{Robust Attributed Network Embedding Preserving Community Information}. In \bibinfo{booktitle}{\emph{ICDE}}. \bibinfo{pages}{1874--1886}.
\newblock


\bibitem[Ma and Zhan(2023)]%
        {SCGDN}
\bibfield{author}{\bibinfo{person}{Yixuan Ma} {and} \bibinfo{person}{Kun Zhan}.} \bibinfo{year}{2023}\natexlab{}.
\newblock \showarticletitle{Self-Contrastive Graph Diffusion Network}. In \bibinfo{booktitle}{\emph{ACM MM}}. \bibinfo{pages}{3857–3865}.
\newblock


\bibitem[MacQueen(1967)]%
        {kmeans}
\bibfield{author}{\bibinfo{person}{J. MacQueen}.} \bibinfo{year}{1967}\natexlab{}.
\newblock \showarticletitle{Some methods for classification and analysis of multivariate observations}.
\newblock


\bibitem[{Moradi} et~al\mbox{.}(2015)]%
        {Moradi2015effective}
\bibfield{author}{\bibinfo{person}{Parham {Moradi}}, \bibinfo{person}{Sajad {Ahmadian}}, {and} \bibinfo{person}{Fardin {Akhlaghian}}.} \bibinfo{year}{2015}\natexlab{}.
\newblock \showarticletitle{{An effective trust-based recommendation method using a novel graph clustering algorithm}}.
\newblock \bibinfo{journal}{\emph{Physica A Statistical Mechanics and its Applications}}  \bibinfo{volume}{436} (\bibinfo{year}{2015}), \bibinfo{pages}{462--481}.
\newblock


\bibitem[Newman(2006a)]%
        {newman2006modularity}
\bibfield{author}{\bibinfo{person}{Mark~EJ Newman}.} \bibinfo{year}{2006}\natexlab{a}.
\newblock \showarticletitle{Modularity and community structure in networks}.
\newblock \bibinfo{journal}{\emph{Proceedings of the national academy of sciences}} \bibinfo{volume}{103}, \bibinfo{number}{23} (\bibinfo{year}{2006}), \bibinfo{pages}{8577--8582}.
\newblock


\bibitem[Newman(2006b)]%
        {Newman2006Finding}
\bibfield{author}{\bibinfo{person}{M.~E.~J. Newman}.} \bibinfo{year}{2006}\natexlab{b}.
\newblock \showarticletitle{Finding community structure in networks using the eigenvectors of matrices}.
\newblock \bibinfo{journal}{\emph{Phys. Rev. E}} (\bibinfo{date}{Sep} \bibinfo{year}{2006}), \bibinfo{pages}{036104}.
\newblock


\bibitem[Pan et~al\mbox{.}(2020)]%
        {ARVGA}
\bibfield{author}{\bibinfo{person}{Shirui Pan}, \bibinfo{person}{Ruiqi Hu}, \bibinfo{person}{Sai-Fu Fung}, \bibinfo{person}{Guodong Long}, \bibinfo{person}{Jing Jiang}, {and} \bibinfo{person}{Chengqi Zhang}.} \bibinfo{year}{2020}\natexlab{}.
\newblock \showarticletitle{Learning Graph Embedding With Adversarial Training Methods}.
\newblock \bibinfo{journal}{\emph{IEEE Transactions on Cybernetics}} \bibinfo{volume}{50}, \bibinfo{number}{6} (\bibinfo{year}{2020}), \bibinfo{pages}{2475--2487}.
\newblock


\bibitem[Park et~al\mbox{.}(2019)]%
        {GALA}
\bibfield{author}{\bibinfo{person}{Jiwoong Park}, \bibinfo{person}{Minsik Lee}, \bibinfo{person}{Hyung~Jin Chang}, \bibinfo{person}{Kyuewang Lee}, {and} \bibinfo{person}{Jin~Young Choi}.} \bibinfo{year}{2019}\natexlab{}.
\newblock \showarticletitle{Symmetric Graph Convolutional Autoencoder for Unsupervised Graph Representation Learning}. In \bibinfo{booktitle}{\emph{ICCV}}. \bibinfo{pages}{6518--6527}.
\newblock


\bibitem[Parulekar et~al\mbox{.}(2023)]%
        {Parulekar2023InfoNCELP}
\bibfield{author}{\bibinfo{person}{Advait Parulekar}, \bibinfo{person}{Liam Collins}, \bibinfo{person}{Karthikeyan Shanmugam}, \bibinfo{person}{Aryan Mokhtari}, {and} \bibinfo{person}{Sanjay Shakkottai}.} \bibinfo{year}{2023}\natexlab{}.
\newblock \showarticletitle{InfoNCE Loss Provably Learns Cluster-Preserving Representations}. In \bibinfo{booktitle}{\emph{Annual Conference Computational Learning Theory}}.
\newblock


\bibitem[Perozzi et~al\mbox{.}(2014)]%
        {deepwalk}
\bibfield{author}{\bibinfo{person}{Bryan Perozzi}, \bibinfo{person}{Rami Al-Rfou}, {and} \bibinfo{person}{Steven Skiena}.} \bibinfo{year}{2014}\natexlab{}.
\newblock \showarticletitle{DeepWalk: Online Learning of Social Representations}. In \bibinfo{booktitle}{\emph{KDD}}. \bibinfo{pages}{701--710}.
\newblock


\bibitem[Salha-Galvan et~al\mbox{.}(2022)]%
        {Guillaume2022Modularity}
\bibfield{author}{\bibinfo{person}{Guillaume Salha-Galvan}, \bibinfo{person}{Johannes~F. Lutzeyer}, \bibinfo{person}{George Dasoulas}, \bibinfo{person}{Romain Hennequin}, {and} \bibinfo{person}{Michalis Vazirgiannis}.} \bibinfo{year}{2022}\natexlab{}.
\newblock \showarticletitle{Modularity-aware graph autoencoders for joint community detection and link prediction}.
\newblock \bibinfo{journal}{\emph{Neural Networks}}  \bibinfo{volume}{153} (\bibinfo{year}{2022}), \bibinfo{pages}{474--495}.
\newblock


\bibitem[Shchur et~al\mbox{.}(2018)]%
        {Oleksandr2018Pitfalls}
\bibfield{author}{\bibinfo{person}{Oleksandr Shchur}, \bibinfo{person}{Maximilian Mumme}, \bibinfo{person}{Aleksandar Bojchevski}, {and} \bibinfo{person}{Stephan G{\"u}nnemann}.} \bibinfo{year}{2018}\natexlab{}.
\newblock \showarticletitle{Pitfalls of Graph Neural Network Evaluation}.
\newblock \bibinfo{journal}{\emph{ArXiv}}  \bibinfo{volume}{abs/1811.05868} (\bibinfo{year}{2018}).
\newblock


\bibitem[Shi and Malik(2000)]%
        {Jianbo2000Normalized}
\bibfield{author}{\bibinfo{person}{Jianbo Shi} {and} \bibinfo{person}{J. Malik}.} \bibinfo{year}{2000}\natexlab{}.
\newblock \showarticletitle{Normalized cuts and image segmentation}.
\newblock \bibinfo{journal}{\emph{IEEE Transactions on Pattern Analysis and Machine Intelligence}} \bibinfo{volume}{22}, \bibinfo{number}{8} (\bibinfo{year}{2000}), \bibinfo{pages}{888--905}.
\newblock


\bibitem[Sun et~al\mbox{.}(2020)]%
        {Sun2020Network}
\bibfield{author}{\bibinfo{person}{Heli Sun}, \bibinfo{person}{Fang He}, \bibinfo{person}{Jianbin Huang}, \bibinfo{person}{Yizhou Sun}, \bibinfo{person}{Yang Li}, \bibinfo{person}{Chenyu Wang}, \bibinfo{person}{Liang He}, \bibinfo{person}{Zhongbin Sun}, {and} \bibinfo{person}{Xiaolin Jia}.} \bibinfo{year}{2020}\natexlab{}.
\newblock \showarticletitle{Network Embedding for Community Detection in Attributed Networks}.
\newblock \bibinfo{journal}{\emph{ACM Trans. Knowl. Discov. Data}} \bibinfo{volume}{14}, \bibinfo{number}{3}, Article \bibinfo{articleno}{36} (\bibinfo{date}{may} \bibinfo{year}{2020}), \bibinfo{numpages}{25}~pages.
\newblock


\bibitem[Sun et~al\mbox{.}(2022)]%
        {Sun2022Graph}
\bibfield{author}{\bibinfo{person}{Jianyong Sun}, \bibinfo{person}{Wei Zheng}, \bibinfo{person}{Qingfu Zhang}, {and} \bibinfo{person}{Zongben Xu}.} \bibinfo{year}{2022}\natexlab{}.
\newblock \showarticletitle{Graph Neural Network Encoding for Community Detection in Attribute Networks}.
\newblock \bibinfo{journal}{\emph{IEEE Transactions on Cybernetics}} \bibinfo{volume}{52}, \bibinfo{number}{8} (\bibinfo{year}{2022}), \bibinfo{pages}{7791--7804}.
\newblock


\bibitem[Sun et~al\mbox{.}(2024)]%
        {SunLCWBZ24}
\bibfield{author}{\bibinfo{person}{Wangbin Sun}, \bibinfo{person}{Jintang Li}, \bibinfo{person}{Liang Chen}, \bibinfo{person}{Bingzhe Wu}, \bibinfo{person}{Yatao Bian}, {and} \bibinfo{person}{Zibin Zheng}.} \bibinfo{year}{2024}\natexlab{}.
\newblock \showarticletitle{Rethinking and Simplifying Bootstrapped Graph Latents}. In \bibinfo{booktitle}{\emph{{WSDM}}}. \bibinfo{publisher}{{ACM}}, \bibinfo{pages}{665--673}.
\newblock


\bibitem[Thakoor et~al\mbox{.}(2021)]%
        {BGRL2021ICLRWorkshop}
\bibfield{author}{\bibinfo{person}{Shantanu Thakoor}, \bibinfo{person}{Corentin Tallec}, \bibinfo{person}{Mohammad~Gheshlaghi Azar}, \bibinfo{person}{Remi Munos}, \bibinfo{person}{Petar Veli{\v{c}}kovi{\'c}}, {and} \bibinfo{person}{Michal Valko}.} \bibinfo{year}{2021}\natexlab{}.
\newblock \showarticletitle{Bootstrapped Representation Learning on Graphs}. In \bibinfo{booktitle}{\emph{ICLR 2021 Workshop on Geometrical and Topological Representation Learning}}.
\newblock


\bibitem[Traag et~al\mbox{.}(2019)]%
        {Leiden}
\bibfield{author}{\bibinfo{person}{V. Traag}, \bibinfo{person}{L. Waltman}, {and} \bibinfo{person}{Nees~Jan van Eck}.} \bibinfo{year}{2019}\natexlab{}.
\newblock \showarticletitle{From Louvain to Leiden: guaranteeing well-connected communities}.
\newblock \bibinfo{journal}{\emph{Scientific Reports}}  \bibinfo{volume}{9} (\bibinfo{date}{03} \bibinfo{year}{2019}), \bibinfo{pages}{5233}.
\newblock


\bibitem[Tsitsulin et~al\mbox{.}(2023)]%
        {DMoN2023JMLR}
\bibfield{author}{\bibinfo{person}{Anton Tsitsulin}, \bibinfo{person}{John Palowitch}, \bibinfo{person}{Bryan Perozzi}, {and} \bibinfo{person}{Emmanuel M{\"{u}}ller}.} \bibinfo{year}{2023}\natexlab{}.
\newblock \showarticletitle{Graph Clustering with Graph Neural Networks}.
\newblock \bibinfo{journal}{\emph{J. Mach. Learn. Res.}}  \bibinfo{volume}{24} (\bibinfo{year}{2023}), \bibinfo{pages}{127:1--127:21}.
\newblock


\bibitem[van~der Maaten and Hinton(2008)]%
        {tsne}
\bibfield{author}{\bibinfo{person}{Laurens van~der Maaten} {and} \bibinfo{person}{Geoffrey Hinton}.} \bibinfo{year}{2008}\natexlab{}.
\newblock \showarticletitle{Visualizing Data using t-SNE}.
\newblock \bibinfo{journal}{\emph{JMLR}} \bibinfo{volume}{9}, \bibinfo{number}{86} (\bibinfo{year}{2008}), \bibinfo{pages}{2579--2605}.
\newblock


\bibitem[Veličković et~al\mbox{.}(2018)]%
        {GAT}
\bibfield{author}{\bibinfo{person}{Petar Veličković}, \bibinfo{person}{Guillem Cucurull}, \bibinfo{person}{Arantxa Casanova}, \bibinfo{person}{Adriana Romero}, \bibinfo{person}{Pietro Liò}, {and} \bibinfo{person}{Yoshua Bengio}.} \bibinfo{year}{2018}\natexlab{}.
\newblock \showarticletitle{Graph Attention Networks}. In \bibinfo{booktitle}{\emph{ICLR}}.
\newblock


\bibitem[Veličković et~al\mbox{.}(2019)]%
        {DGI}
\bibfield{author}{\bibinfo{person}{Petar Veličković}, \bibinfo{person}{William Fedus}, \bibinfo{person}{William~L. Hamilton}, \bibinfo{person}{Pietro Liò}, \bibinfo{person}{Yoshua Bengio}, {and} \bibinfo{person}{R~Devon Hjelm}.} \bibinfo{year}{2019}\natexlab{}.
\newblock \showarticletitle{Deep Graph Infomax}. In \bibinfo{booktitle}{\emph{ICLR}}.
\newblock


\bibitem[Wang and Liu(2021)]%
        {Wang2021Understanding}
\bibfield{author}{\bibinfo{person}{Feng Wang} {and} \bibinfo{person}{Huaping Liu}.} \bibinfo{year}{2021}\natexlab{}.
\newblock \showarticletitle{Understanding the Behaviour of Contrastive Loss}. In \bibinfo{booktitle}{\emph{CVPR}}. \bibinfo{pages}{2495--2504}.
\newblock


\bibitem[Wu et~al\mbox{.}(2018)]%
        {Wu2018CVPRUnsupervised}
\bibfield{author}{\bibinfo{person}{Zhirong Wu}, \bibinfo{person}{Yuanjun Xiong}, \bibinfo{person}{Stella~X. Yu}, {and} \bibinfo{person}{Dahua Lin}.} \bibinfo{year}{2018}\natexlab{}.
\newblock \showarticletitle{Unsupervised Feature Learning via Non-parametric Instance Discrimination}. In \bibinfo{booktitle}{\emph{2018 IEEE/CVF Conference on Computer Vision and Pattern Recognition}}. \bibinfo{pages}{3733--3742}.
\newblock


\bibitem[Yang et~al\mbox{.}(2016)]%
        {Yang2016Modularity}
\bibfield{author}{\bibinfo{person}{Liang Yang}, \bibinfo{person}{Xiaochun Cao}, \bibinfo{person}{Dongxiao He}, \bibinfo{person}{Chuan Wang}, \bibinfo{person}{Xiao Wang}, {and} \bibinfo{person}{Weixiong Zhang}.} \bibinfo{year}{2016}\natexlab{}.
\newblock \showarticletitle{Modularity based community detection with deep learning}. In \bibinfo{booktitle}{\emph{IJCAI}} \emph{(\bibinfo{series}{IJCAI'16})}. \bibinfo{pages}{2252–2258}.
\newblock


\bibitem[Yang et~al\mbox{.}(2023)]%
        {CCGC2023AAAI}
\bibfield{author}{\bibinfo{person}{Xihong Yang}, \bibinfo{person}{Yue Liu}, \bibinfo{person}{Sihang Zhou}, \bibinfo{person}{Siwei Wang}, \bibinfo{person}{Wenxuan Tu}, \bibinfo{person}{Qun Zheng}, \bibinfo{person}{Xinwang Liu}, \bibinfo{person}{Liming Fang}, {and} \bibinfo{person}{En Zhu}.} \bibinfo{year}{2023}\natexlab{}.
\newblock \showarticletitle{Cluster-Guided Contrastive Graph Clustering Network}. In \bibinfo{booktitle}{\emph{{AAAI}}}. \bibinfo{publisher}{{AAAI} Press}, \bibinfo{numpages}{9}~pages.
\newblock


\bibitem[You et~al\mbox{.}(2020)]%
        {GraphCL}
\bibfield{author}{\bibinfo{person}{Yuning You}, \bibinfo{person}{Tianlong Chen}, \bibinfo{person}{Yongduo Sui}, \bibinfo{person}{Ting Chen}, \bibinfo{person}{Zhangyang Wang}, {and} \bibinfo{person}{Yang Shen}.} \bibinfo{year}{2020}\natexlab{}.
\newblock \showarticletitle{Graph Contrastive Learning with Augmentations}. In \bibinfo{booktitle}{\emph{NeurIPS}}, Vol.~\bibinfo{volume}{33}. \bibinfo{pages}{5812--5823}.
\newblock


\bibitem[Zhou et~al\mbox{.}(2022)]%
        {Zhou2022End}
\bibfield{author}{\bibinfo{person}{Cangqi Zhou}, \bibinfo{person}{Yuxiang Wang}, \bibinfo{person}{Jing Zhang}, \bibinfo{person}{Jiqiong Jiang}, {and} \bibinfo{person}{Dianming Hu}.} \bibinfo{year}{2022}\natexlab{}.
\newblock \showarticletitle{End-to-end Modularity-based Community Co-partition in Bipartite Networks}. In \bibinfo{booktitle}{\emph{CIKM}}. \bibinfo{pages}{2711–2720}.
\newblock


\bibitem[Zhu et~al\mbox{.}(2020)]%
        {GRACE2020ICMLWorkshop}
\bibfield{author}{\bibinfo{person}{Yanqiao Zhu}, \bibinfo{person}{Yichen Xu}, \bibinfo{person}{Feng Yu}, \bibinfo{person}{Qiang Liu}, \bibinfo{person}{Shu Wu}, {and} \bibinfo{person}{Liang Wang}.} \bibinfo{year}{2020}\natexlab{}.
\newblock \showarticletitle{{Deep Graph Contrastive Representation Learning}}. In \bibinfo{booktitle}{\emph{ICML Workshop on Graph Representation Learning and Beyond}}.
\newblock


\end{thebibliography}

\appendix
\clearpage
\balance

\section{Algorithm}
To help better understand the proposed framework, we provide the detailed algorithm for training \ours in Algorithm~\ref{alg:Framwork}.

\begin{algorithm}[h]
  \caption{Community-Aware Graph Clustering (\ours)}
  \label{alg:Framwork}
  \begin{algorithmic}[1]
    \REQUIRE ~~ Graph $\mathcal{G} = (\mathcal{V}, \mathcal{E}, \mathbf{X})$, encoder $f_{\theta}(\cdot)$, Randomly sampled node set $\mathcal{N}$, number of walks $t$, depth of walks $l$, hyperparameter $\alpha$;
    \ENSURE ~~ Learned encoder $f_{\theta}(\cdot)$;
    \WHILE{\textit{not converged}}
    \STATE Perform random walk for each node in $\mathcal{N}$, obtain $\mathcal{B}$;
    \STATE Perform random walk for each node in $\mathcal{B}$, obtain similarity matrix $\mathbf{S}$;
    \STATE Calculate modularity matrix $\mathbf{B}$ according to Eq.(\ref{eq:minibatch_modularity_matrix});
    \STATE Calculate $\mathcal{L}_{SimCLR}$ according to Eq.(\ref{eq:SimCLR_loss});
    \STATE Update $\theta$ by gradient descent;
    \ENDWHILE
    \RETURN $f_{\theta}(\cdot)$;
  \end{algorithmic}
\end{algorithm}

\section{MORE DETAILS ABOUT THE Complexity analysis}
We report the complexity analysis from both theoretical and experimental perspectives in Table~\ref{table:complexity}.

\begin{table}[b]
  \caption{Time and space analyses of different methods. Where $k$ is the number of clusters. The experimental GPU memory costs and time costs (training) are obtained on Cora dataset.}
  \label{table:complexity}
  \begin{adjustbox}{width=0.5\textwidth,center}
    \large 
    \centering
    \begin{tabular}{l|cccc}
      \toprule
      \addlinespace
      \textbf{Method} & \textbf{Time Complexity}        & \textbf{Space Complexity}         & \textbf{Memory Cost(MB)} & \textbf{Time Cost(s)} \\
      \addlinespace
      \midrule
      \midrule

      Node2vec        & $\mathcal{O}(bd)$               & $\mathcal{O}(Nd)$                 & 1,122                    & 62.1                  \\

      DGI             & $\mathcal{O}(md + Nd^2)$        & $\mathcal{O}(m + Nd + d^2)$       & 218                      & 36.8                  \\

      GRACE           & $\mathcal{O}(N^2 d+md+d^2)$     & $\mathcal{O}(m+Nd)$               & 277                      & 2.9                   \\

      BGRL            & $\mathcal{O}(md+Nd^2)$          & $\mathcal{O}(m+Nd+d^2)$           & 1,201                    & 31.2                  \\

      MVGRL           & $\mathcal{O}(N^2 d+Nd^2)$       & $\mathcal{O}(N^2 + Nd + d^2)$     & 2,924                    & 33.9                  \\

      SCGDN           & $\mathcal{O}(N^2 + Nd^2)$       & $\mathcal{O}(N^2 + Nd + d^2)$     & 1,080                    & 31.9                  \\

      CCGC            & $\mathcal{O}(kNd+N^2 d + Nd^2)$ & $\mathcal{O}(N^2 + (N+k)d + d^2)$ & 3,233                    & 38.0                  \\
      DGCLUSTER       & $\mathcal{O}(N^2 d+Nd^2)$       & $\mathcal{O}(N^2 + Nd + d^2)$     & 267                      & 6.5                   \\

      DMoN            & $\mathcal{O}((m+N)k)$           & $\mathcal{O}(m+Nk)$               & 301                      & 9.3                   \\

      S${^3}$GC       & $\mathcal{O}(Nsd^2)$            & $\mathcal{O}(Nd+bsd+d^2)$         & 2,148                    & 27.2                  \\

      \ours           & $\mathcal{O}(Nd^2)$             & $\mathcal{O}(m+Nd+b^2)$           & 209                      & 2.7                   \\
      \bottomrule
    \end{tabular}
  \end{adjustbox}
\end{table}

\section{MORE DETAILS ABOUT THE EXPERIMENTS}
\subsection{Hyperparameter settings}
We report our hyperparameter settings in Table~\ref{tab:hyperparameter}.

\begin{table}[b]
  \centering
  \caption{Hyperparameter settings, where $d$ is the embedding dimension, $\alpha$ is the activation threshold, $\tau$ is the temperature, $t$ is the number of random walks, $l$ is the depth of random walks.}
  \begin{adjustbox}{width=0.5\textwidth,center}
    \large 
    \begin{tabular}{l|cccccccc}
      \toprule
      \textbf{Dataset}         & \#GNN layers & $d$ & $\alpha$ & $\tau$ & $t$ & $l$ & learning rate & weight decay \\
      \midrule
      \textbf{Cora}            & 1            & 512 & 0.5      & 0.3    & 100 & 2   & 0.0005        & 0.001        \\
      \textbf{CiteSeer}        & 2            & 512 & 0.5      & 0.9    & 100 & 3   & 0.0001        & 0.0005       \\
      \textbf{Photo}           & 1            & 512 & 0.5      & 0.5    & 100 & 3   & 0.0005        & 0.001        \\
      \textbf{Computers}       & 2            & 512 & 0.1      & 0.9    & 100 & 3   & 0.0005        & 0.001        \\
      \textbf{ogbn-arxiv}      & 2            & 256 & 0.1      & 0.9    & 20  & 5   & 0.01          & 0            \\
      \textbf{Reddit}          & 2            & 256 & 0.5      & 0.5    & 20  & 5   & 0.01          & 0            \\
      \textbf{ogbn-products}   & 3            & 256 & 0.1      & 0.9    & 20  & 4   & 0.01          & 0            \\
      \textbf{ogbn-papers100M} & 2            & 64  & 0.1      & 0.9    & 20  & 3   & 0.01          & 0            \\
      \bottomrule
    \end{tabular}
    \label{tab:hyperparameter}
  \end{adjustbox}
\end{table}

\subsection{Ablation study on different mechanisms}
We report the ablation study results on different mechanisms in Table~\ref{tab:abltion}.

\begin{table*}[t]
  \caption{Ablations on different components of \ours.}
  \label{tab:abltion}
  \centering
  \begin{tabular}{lcccccc}
    \toprule

    \textbf{Dataset}                                      & \textbf{Metric} & \ours(w/ SL)      & \ours(w/ MS)      & \ours(w/ EI)      & \ours(w/ HMS)                      & \ours             \\

    \midrule
    \addlinespace
    \multirow{4}{*}{\makecell[c]{\textbf{Cora}}}          & ACC             & 0.703             & \underline{0.754} & 0.723             & 0.753                              & \textbf{0.760}    \\
                                                          & NMI             & 0.541             & 0.582             & 0.562             & \underline{0.589}                  & \textbf{0.597}    \\
                                                          & ARI             & 0.492             & 0.559             & 0.518             & \underline{0.561}                  & \textbf{0.573}    \\
                                                          & F1              & 0.687             & \underline{0.736} & 0.713             & 0.729                              & \textbf{0.739}    \\
    \midrule
    \multirow{4}{*}{\makecell[c]{\textbf{CiteSeer}}}      & ACC             & 0.701             & 0.700             & \underline{0.703} & 0.698                              & \textbf{0.706}    \\
                                                          & NMI             & \textbf{0.452}    & 0.446             & 0.447             & \underline{0.448}                  & \textbf{0.452}    \\
                                                          & ARI             & \underline{0.462} & 0.459             & 0.443             & 0.443                              & \textbf{0.468}    \\
                                                          & F1              & 0.645             & \textbf{0.653}    & 0.632             & 0.639                              & \underline{0.648} \\
    \midrule
    \multirow{4}{*}{\makecell[c]{\textbf{Photo}}}         & ACC             & 0.672             & 0.749             & 0.736             & \underline{0.769}                  & \textbf{0.790}    \\
                                                          & NMI             & 0.594             & 0.645             & 0.634             & \underline{0.683}                  & \textbf{0.716}    \\
                                                          & ARI             & 0.473             & 0.562             & 0.541             & \underline{0.576}                  & \textbf{0.615}    \\
                                                          & F1              & 0.684             & \underline{0.715} & 0.713             & 0.706                              & \textbf{0.729}    \\
    \midrule
    \multirow{4}{*}{\makecell[c]{\textbf{Computers}}}     & ACC             & 0.496             & 0.590             & 0.590             & \textbf{0.639}                     & \underline{0.620} \\
                                                          & NMI             & 0.548             & 0.580             & 0.574             & \textbf{0.604}                     & \underline{0.592} \\
                                                          & ARI             & 0.342             & 0.438             & 0.432             & \textbf{0.475}                     & \underline{0.462} \\
                                                          & F1              & 0.395             & 0.558             & 0.543             & \textbf{0.580}                     & \underline{0.574} \\
    \midrule
    \multirow{4}{*}{\makecell[c]{\textbf{ogbn-arxiv}}}    & ACC             & 0.235             & \underline{0.345} & 0.336             & \multirow{4}{*}{\makecell[c]{OOM}} & \textbf{0.388}    \\
                                                          & NMI             & 0.265             & \underline{0.463} & 0.457             &                                    & \textbf{0.469}    \\
                                                          & ARI             & 0.227             & \underline{0.252} & 0.247             &                                    & \textbf{0.310}    \\
                                                          & F1              & 0.089             & \textbf{0.270}    & 0.251             &                                    & \underline{0.266} \\
    \midrule
    \multirow{4}{*}{\makecell[c]{\textbf{Reddit}}}        & ACC             & 0.240             & \underline{0.832} & 0.785             & \multirow{4}{*}{\makecell[c]{OOM}} & \textbf{0.911}    \\
                                                          & NMI             & 0.360             & \underline{0.852} & 0.836             &                                    & \textbf{0.875}    \\
                                                          & ARI             & 0.145             & \underline{0.842} & 0.783             &                                    & \textbf{0.907}    \\
                                                          & F1              & 0.134             & \underline{0.714} & 0.660             &                                    & \textbf{0.853}    \\
    \midrule

    \multirow{4}{*}{\makecell[c]{\textbf{ogbn-products}}} & ACC             & 0.144             & \underline{0.391} & 0.387             & \multirow{4}{*}{\makecell[c]{OOM}} & \textbf{0.425}    \\
                                                          & NMI             & 0.079             & \underline{0.525} & 0.510             &                                    & \textbf{0.551}    \\
                                                          & ARI             & 0.059             & \underline{0.192} & 0.177             &                                    & \textbf{0.215}    \\
                                                          & F1              & 0.035             & \underline{0.244} & 0.222             &                                    & \textbf{0.276}    \\

    \bottomrule
  \end{tabular}
\end{table*}

\end{document}